\RequirePackage{amsmath} % To avoid the `unable to redefine vec' warning
\documentclass[smallextended]{svjour3}       % onecolumn (second format)
\usepackage[utf8]{inputenc}

\usepackage{amsfonts}

\usepackage{amsmath}

\usepackage{amssymb}

\usepackage{blindtext}

\usepackage{booktabs}

\usepackage{subcaption}

\usepackage[usenames,dvipsnames]{color}

\usepackage[strict]{changepage}

\usepackage{dsfont}

\usepackage{enumitem}

\usepackage{framed}

\usepackage{geometry}
\usepackage[normalem]{ulem}

\usepackage{wrapfig}

\usepackage{placeins}

\usepackage[numbers]{natbib}

\usepackage{indentfirst}

\usepackage{mathabx}

\usepackage{mathrsfs}

\usepackage{multirow}

\usepackage{natbib}

\usepackage{soul}
    \soulregister\cite7
    \soulregister\ref7
    \soulregister\pageref7

\usepackage{truncate}

\usepackage{verbatim}

\usepackage{hyperref}
    \hypersetup{% Hyperref package: avoid coloured border
        pdfborder = {0 0 0}%
    }%
    
\usepackage[T1, OT1]{fontenc}
    \DeclareTextSymbolDefault{\dh}{T1}  % Use the symbol \dh

\usepackage{pdflscape}

\usepackage{afterpage}

\usepackage{forest}

\usepackage{xparse}
\usepackage{todonotes}
    \presetkeys{todonotes}{inline}{}
    \NewDocumentCommand{\task}{o m}{\IfValueTF{#1}{%
        \todo[color=yellow!80]{\texttt{[#1]} #2}%
    }{
        \todo[color=yellow!80]{#2}%
    }}

\usepackage{listings}
\usepackage{algpseudocode}
    \definecolor{formalshade}{rgb}{0.95,0.95,1}
    
    \newenvironment{alg-notitle}[1][Algorithm]{% 1 parameter: description of the function.
      \MakeFramed{\advance\hsize-\width\FrameRestore}%
      \noindent\hspace{-4.55pt}% disable indenting first paragraph
      \begin{adjustwidth}{}{7pt}%
      \vspace{-12pt}%
      {\large\textbf{#1}}
      \vspace{0.3em}
      \begin{algorithmic} %number every line
    }
    {%
      \end{algorithmic}%
      \vspace{2pt}%
      \end{adjustwidth}%
      \endMakeFramed%
    }
    \algrenewcommand{\algorithmiccomment}[1]{\hskip3em// #1}
    \algblockx[funct]{Function}{EndFunction}
       [2][Function]{\textbf{#1} (#2)}
       [1][~]{\textbf{return} #1}

\newcommand{\revisionEnabled}{%
    \definecolor{shadecolor}{HTML}{FFE27E}%
    \definecolor{redish}{HTML}{FF2222}%
    \renewcommand{\revisionAdded}[1]{%
        \ignorespaces%
        \sethlcolor{shadecolor}%
        \hl{##1}%
        \unskip%
    }%
    \renewcommand{\revisionDeleted}[1]{%
        \ignorespaces%
        \setstcolor{redish}%
        \st{##1}%
        \unskip%
    }%
    \renewcommand{\revisionChanged}[2]{%
        \revisionDeleted{##1}%
        {\ }% Blank space
        \revisionAdded{##2}%
    }%
    \renewcommand{\nextReviewer}[1]{%
        \stepcounter{ReviewerNumber} % Add 1 to counter
        \setcounter{CommentNumber}{0}%
        \vspace*{0.4cm} \noindent%
        \textbf{\large Reviewer \#\theReviewerNumber}%
    }%
    \renewcommand{\revisionComment}[2]{%
        \stepcounter{CommentNumber} % Add 1 to counter
        \begin{center}%
            \begin{minipage}{.9\linewidth}%
              \begin{shaded}%
                \sffamily\slshape%
                Reviewer \#\theReviewerNumber.
                Comment \theCommentNumber. 
                ##1%
              \end{shaded}%
            \end{minipage}%
          \end{center}%
        \ignorespaces%
        ##2%
        \unskip%
        \vspace{0.3cm}
    }%
}%
\newcommand{\revisionAdded}[1]{}%
\newcommand{\revisionDeleted}[1]{}%
\newcommand{\revisionChanged}[2]{}%
\newcommand{\nextReviewer}[2]{}%
\newcommand{\revisionComment}[2]{}%
\newcounter{CommentNumber}%
\newcounter{ReviewerNumber}%
\revisionEnabled%
\let\oldCopyright\copyright
\renewcommand{\copyright}{\textsuperscript{\oldCopyright}}

\newcommand{\algorithmname}[1]{#1\xspace}
\newcommand{\watershed}{\algorithmname{watershed}}

\newcommand{\Houghtransform}{\algorithmname{Hough transform}}

\newcommand{\SVM}{\algorithmname{SVM}}
\newcommand{\RandomForest}{\algorithmname{random forest}}
\newcommand{\GradientBoosting}{\algorithmname{gradient boosting}}

\newcommand{\MultilayerPerceptron}{\algorithmname{multi-layer perceptron}}

\newcommand{\colorspacename}[1]{\text{#1}\xspace}
\newcommand{\channelname}[1]{\colorspacename{#1}}
\newcommand{\RGB}{\colorspacename{RGB}}
\newcommand{\HSV}{\colorspacename{HSV}}
\newcommand{\CIELab}{\colorspacename{CIELAB}}

\newcommand{\Fmeasure}{$\text{F}_1$-measure}
\newcommand{\Sensitivity}{Sensitivity}
\newcommand{\Specificity}{Specificity}
\newcommand{\Precision}{Precision}
\newcommand{\Accuracy}{Accuracy}
\newcommand{\CohensKappa}{Cohen's $\kappa$}

\smartqed  % flush right qed marks, e.g. at end of proof
\journalname{Multimedia Tools and Applications}

\begin{document}
% (Computer Vision)
\title{Geometric-Based Nail Segmentation for Clinical Measurements}


\author{Bernat Galm\'{e}s   \and
        Gabriel Moy\`{a}-Alcover % https://orcid.org/0000-0002-3412-5499
        \and
         Pedro Bibiloni % https://orcid.org/0000-0002-1825-009 
        \and 
        Javier Varona % https://orcid.org/0000-0002-0287-0486 
        \and
        Antoni Jaume-i-Cap\'{o} %https://orcid.org/0000-0003-3312-5347
        }

\authorrunning{B. Galm\'{e}s \textit{et al.}} % if too long for running head

\institute{B. Galm\'{e}s \and G. Moy\`{a}-Alcover \and P. Bibiloni \and Antoni Jaume-i-Cap\'{o} \and Javier Varona     \at
              University of the Balearic Islands, Dpt. of Mathematics and Computer Science, 07122 Palma, Spain \\
              \email{gabriel.moya@uib.es} \\
              Phone: +34971171399 \\
           \and
           P. Bibiloni \at
              Balearic Islands Health Research Institute (IdISBa), 07010 Palma, Spain.
}


\maketitle
\begin{abstract}

% ABSTRACT
%%% 150-250 words
%
% 1-line summary
%\sout{We introduce a method to segment the nail of the big toe in photographs captured with a template specially designed for this task.
% Motivation
%It is currently used in assessing the evolution of the area affected by a specific pathology in a medical study.}
We present a robust segmentation algorithm that can be used to obtain measurements of toe nails. The presented method assists in a medical study to objectively quantify the incidence of a specific pathology.
Towards such assessment, we require a segmentation of the nail, which locally appears to be very similar to skin.
% How we do it
Several algorithms are used, each of them leveraging a different aspect of the toenail appearance.
We use the Hough transform to locate the tip of the toe and to estimate the nail location and size.
Then, we classify the image's super-pixels based on their geometry and photometric information. 
Afterwards, the watershed transform delineates the border of the nail.
% What results we obtain
The method has been validated with a 348-image medical dataset, achieving an accuracy of 0.993 and an F-measure of 0.925.
%The overall shape of the nail is successfully captured
%Several performance measures are employed to quantify the ability of the method to discriminate nail and toe pixels.
%In terms of performance measures, it achieves an accuracy of 0.993 and an F-measure of 0.925.
The method is robust across samples unalike in nail shape, skin pigmentation, illumination conditions and, more importantly, under the appearance of regions affected by a medical condition.

\end{abstract}

\section{Introduction}
\label{sec:Introduction}

Human nails, mainly made of keratin, present a color very similar to that of the skin.
Both regions have, typically, a somewhat flat texture.
Although their shape tends to be circular, there is a wide diversity of them, regarding their eccentricity and shape of contours.
Thus, very few assumptions can be made to reliably characterize nails in terms of their visual appearance.

%\subsection{Related work}

The problem of toenail segmentation has \emph{not} been widely studied in the literature.To our knowledge, there is not a single research project whose goal has been exclusively to segment the toenail. 

However, this task is tightly related to segmenting the fingernails in images. Most of the existing research is focused on biometric systems and disease detection. See the work of Mente and Marulkar \cite{mente2017review} for a review on various research works based on fingernail disease detection using image analysis. There is also a review on nail image processing for disease detection by Juna V. V. and Dinil Dhananjayan \cite{juna2019review}. 

Most research focuses on segmenting the nail from images captured in environments with a controlled perspective and constant lighting. Under these settings, simple methods can be employed, such as considering color difference between the nail and the hand.

%Some approaches are based on the difference of color between fingers and nails, implicitly assuming a specific illumination and skin tone. 
Fukunaka \emph{et al.} segment fingernails in the \HSV colorspace by thresholding using experimental values \cite{fukunaka2012recognition}. Tolentino \emph{et al.} \cite{tolentino2018detection} uses the same approach.

Fujishima and Hoshino propose a fingernail detection method using the distribution density of nail and finger regions \cite{fujishima2012fingernail}, 
and further improve their method's accuracy by considering the color continuity \cite{fujishima2013fingernail}. 

Gauns Dessai and Borkar \cite{gaunsdessai} segment the nail colors using L*a*b* color space and $k$-means clustering. A similar approach is used by Marulkar and Mente in  \cite{marulkar2018nail}  where they applied  $k$-means clustering, L*a*b* colorspace and Marker-Controlled Watershed Segmentation for better accuracy and precision, although was only applied to a single image.

Wang \emph{et al.} also perform hand segmentation and fingertip detection using color information \cite{wang2013efficient}. They also design a color space for this specific task.In order to segment the fingernail, they use a pixel-wise classifier based on the different distribution of the channel $U_\text{skin}$ in skin and fingernail regions.

The method proposed by Kumar \textit{et al.} \cite{kumar2014biometric} first finds the nail ROI of the index, middle and ring fingers from the hand image. They then segment it by applying a fixed threshold in a grayscale image, and finally refine the grown nail plate with a Gabor filter.

Lee \emph{et al.}  propose  \cite{lee2017image} an image preprocessing method to segment different parts of nail: the lunula and the nail plate. In order to maintain the nail image quality, this paper uses microscopic imagery. %% SUPERSIMPLE

Also working in a controlled scenario, Easwaramoorthy \textit{et al.} \cite{easwaramoorthy16} identify the difficulty to segment the nail bed since the edges in the nail images are not continuous. They proposed an algorithm to extract a set of nail semantic points instead of segmenting the nail itself.

Other approaches allow more flexibility on the input images' illumination, but still control the image perspective, typically because these are approaches designed to be used in biometrics.

To achieve illumination invariance, Barbosa \emph{et al.}~\cite{barbosa2013transient} proposed a nail segmentation algorithm using an active shape model employing local binary pattern features. They also introduced a dataset with controlled captured conditions. %\textcolor{red}{Same authors \cite{barbosa2016use} uses an object detector to obtain a ROI that contains the fingernail.}

Kumuda \emph{et al.} \cite{kumuda2015human} used the \watershed algorithm on a contrast-enhaced grayscale image to segment the distal region of the nail, the nail plate and the finger. 
Later on, the same authors  \cite{kumuda2016characterization} considered an illumination correction stage and an iterative histogram-based thresholding in each component of the \RGB\ color space to binarize the fingernail, whose shape based classified as either oval, round, or rectangular.

Kurniastuti presented a method that uses an active shape model \cite{kurniastuti2018active}. It consists of three steps: grayscaling; contrast stretching, to repair contrast in the image; and use of active shape model. The method, which segments the fingernail area, needs 45 minutes for every sample.

The solutions proposed in the state-of-the-art are designed for high controlled environments, where they can decide the light direction, its intensity. They can also control the image perspective and the cameras used to take the image.
%---------------------------------------------

In this paper, we present a robust toenail segmentation algorithm that can be used for measuring the area of the nail of the big toe in human patients to assist in a medical study towards the objective quantification of the incidence of a specific pathology. To ensure the correctness of the quantification, it is important to segment the toenail correctly from images taken from different cameras, angles and  lighting conditions. Once the nail is segmented, its area is computed and the nail region can be further processed by researchers.

The importance of this method relies on providing a quantitative result that is free from the slight differences that practitioners might introduce if they perform this specific task.  

In this study, made possible by a joint effort in which 348  samples were collected during the clinical practice, we aim at segmenting the toenail. This is a tedious and repetitive task, whose automation releases medical practitioners from this time-consuming effort.

Although it is a simple task for humans, automatically detecting nails in images of human toes is a challenging problem. Computer vision techniques can automate and standardize a well-defined task.

We emphasize that, as of today, the nail segmentation algorithm presented here is being used as a first step in an ongoing medical study.

The objectives initially proposed for such algorithm are:
\begin{enumerate}
    \item Design and implement an algorithm that, given a photograph of the big toe in a especially-designed template, is able to segment the nail region.
    \item Measure its performance compared to human-provided segmentation.
    \item Iterate over the previous steps to reach an algorithm whose results are comparable to human performance.
\end{enumerate}

%\subsection{Original}

%Automatic vision methods play an important role in the medical industry. 
%They provide important insights in numerous applications \cite{walczak2018toward, strisciuglio2016supervised, hesar2018multiple}.
%For instance, to automatize tedious and repetitive tasks that have been traditionally performed by humans. 

%In this paper, we present a robust toenail segmentation algorithm for measuring the area of the nail of the big toe in human patients. 
The algorithm presented here is proposed to assist in a medical study towards the objective quantification of the incidence of a specific pathology.

\subsection{Structure of the paper}

The structure of the rest of the paper is as follows.
In Sect. \ref{sec:ToenailDetection} we introduce the method for segmenting nails.
We cover all steps in detail: location of the fingertip and nail using the \algorithmname{Hough transform}{}; super-pixel creation with quickshift, based on color similarity; subsequent classification with \GradientBoosting; and a \watershed-based refinement to provide the final nail mask.
Sect. \ref{sec:Results} exhibits the behaviour of our method, including an analysis of its performance measured with quantitative metrics.
We conclude, in Sect. \ref{sec:Conclusion}, by studying the strengths and shortcomings of our proposed method.

\section{Toenail Segmentation Method}
\label{sec:ToenailDetection}

The nail segmentation method is composed of several steps: tip of the toe location, nail circle estimation, super-pixel classification and, finally, toenail pixel-wise segmentation. Each of these steps refines the result provided by the previous one.
We start describing the characteristics of the images we employed which define, implicitly, the problem we are facing.

\subsection{High-level Description of Toenails}
\label{sec:ToenailDetection:Problem}

As we stated in the introduction, toenails are part of the outer layer of the skin.
They are located at the end of the toe and have a slightly different color from the skin due to its composition: a hardened or horny cutaneous structure formed of keratin. 
Nails are composed of two parts: 
the lunula, a lighter region from which the nail grows;
the nail plate, which covers the central part of the nail;
and the distal end, in which the nail is no longer attached to the finger or toe surface.
Nails are usually circular, although there is a wide diversity in terms of shape, being some of them more similar to an ellipse or having squared corners.

\begin{figure}[!ht]
    \centering
    \begin{subfigure}[t]{0.3\textwidth}
            \centering
            \includegraphics[width=\textwidth]{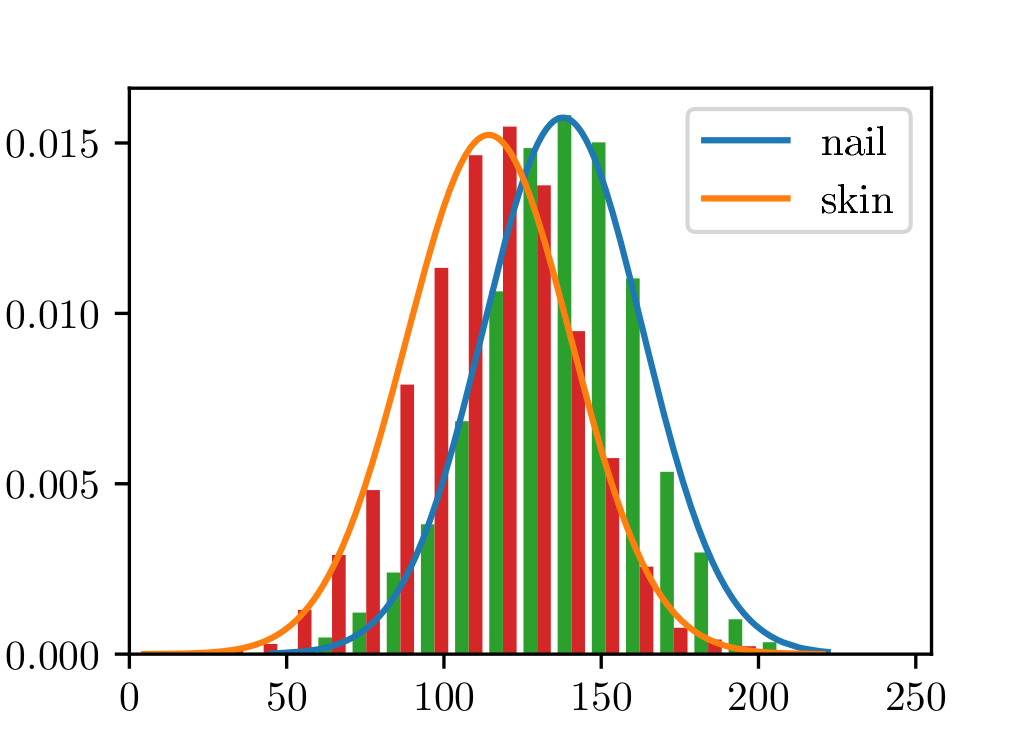}
            \caption{L channel.}
            \label{fig:lab_dist_L}
    \end{subfigure}
    \hfill %add desired spacing between images, e. g. ~, \quad, \qquad etc.
    \begin{subfigure}[t]{0.3\textwidth}
            \centering
            \includegraphics[width=\textwidth]{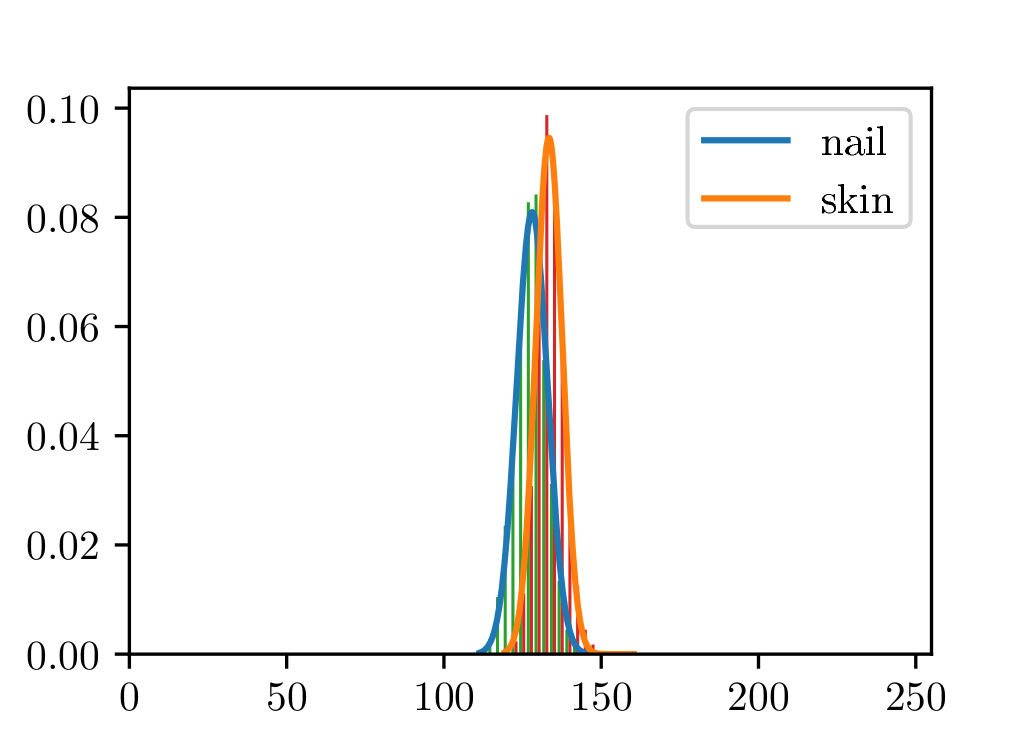}
            \caption{\textit{a} channel.}
            \label{fig:lab_dist_a}
    \end{subfigure}
    \hfill %add desired spacing between images, e. g. ~, \quad, \qquad etc.
    \begin{subfigure}[t]{0.3\textwidth}
            \centering
            \includegraphics[width=\textwidth]{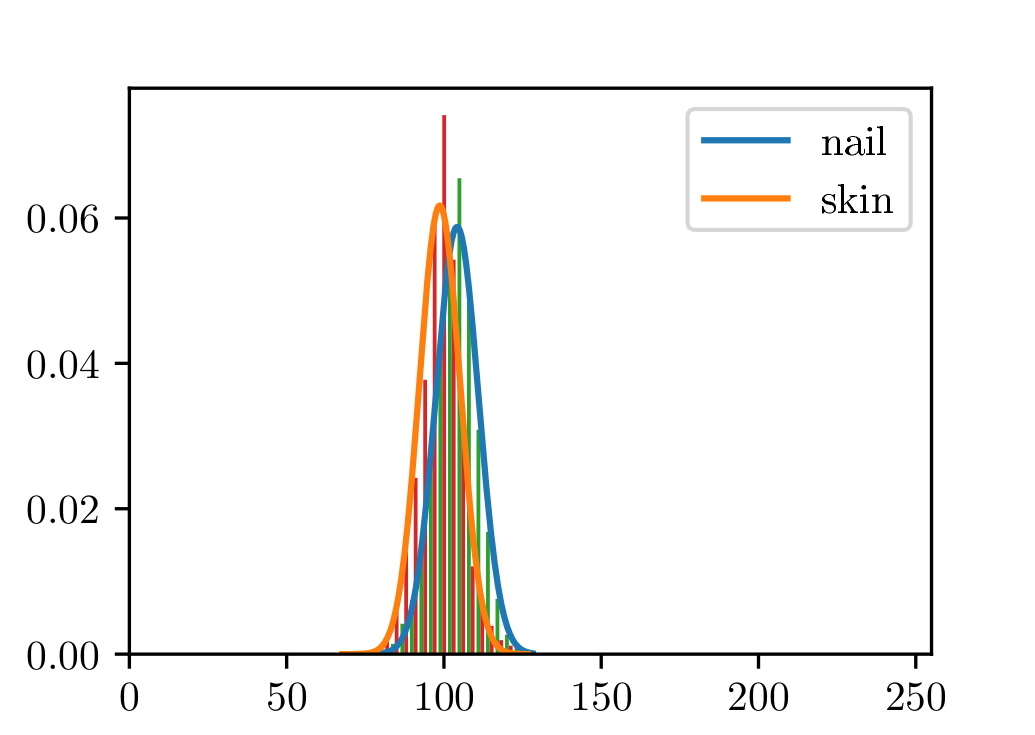}
            \caption{\textit{b} channel.}
            \label{fig:lab_dist_b}
    \end{subfigure}
    \caption{Distribution of nail and toe pixel values accross the three channels of the \CIELab color space.}
    \label{fig:photometric}
\end{figure}

The photometric properties of nail pixels do not contain enough information to segment them from toe ones.
In Fig. \ref{fig:photometric} we group pixels belonging to the nail region and the toe region, respectively.
We observe that their distribution across different channels of the \CIELab color space is very similar.
In fact, none of the channels, nor a combination of them, has proved to be enough to tell apart these pixels, especially when considering pictures taken under different illumination conditions.
Thus, although nail and toe can be easily discriminated by a human observer, doing so based only on pixel-based local information poses as an arduous challenge.

\begin{figure*}[ht!]
    \centering
    \includegraphics[width=0.7\textwidth]{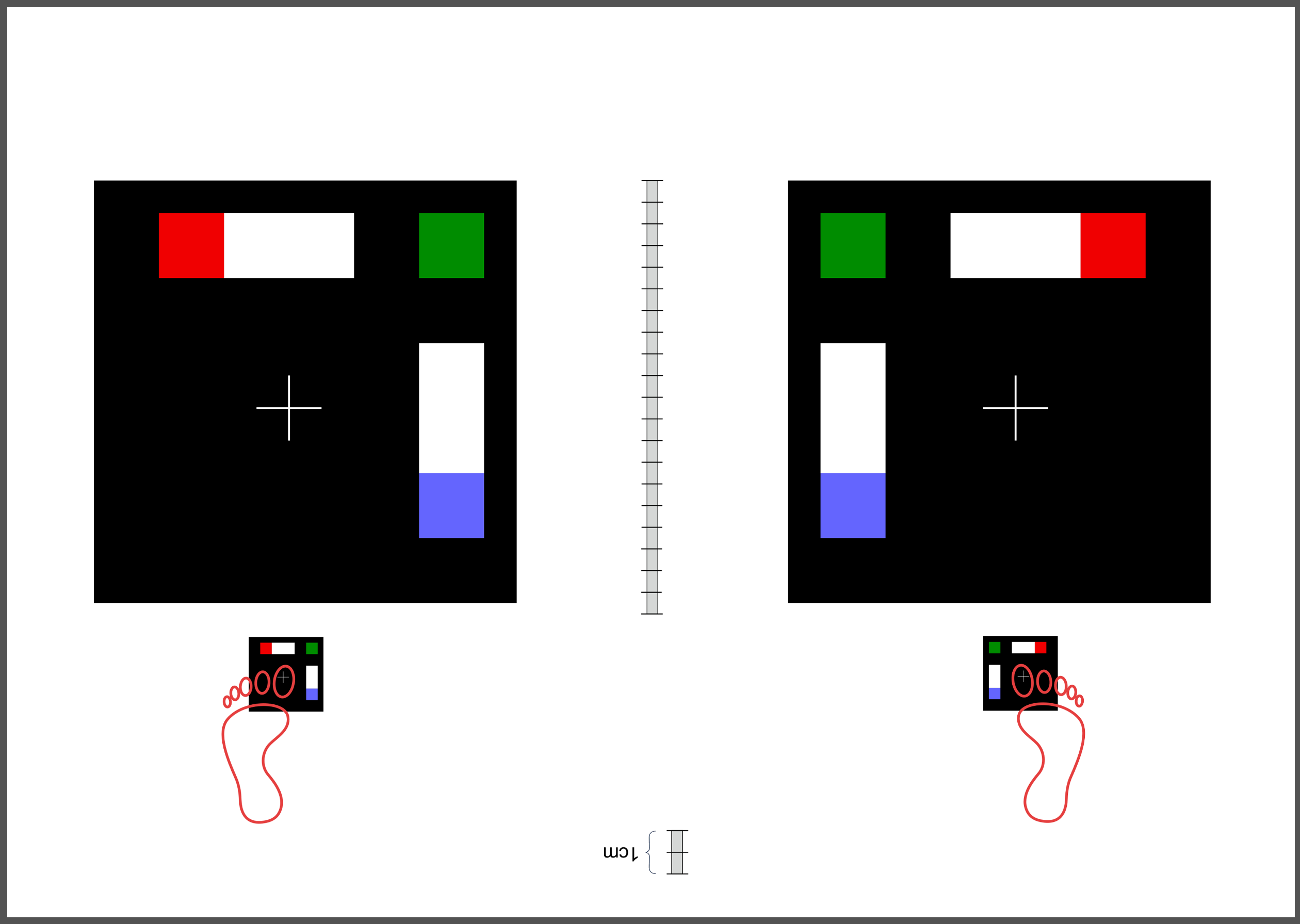}
    \caption{Designed template for both feet, to be printed in a high-quality matte paper of size A4.}
    \label{fig:plantilla}
\end{figure*}

The dataset used in this study was taken in the clinical practice.
Following the guidelines of a well-defined medical study, images were acquired in real conditions.
All of them were captured in the doctor's office with their smartphones' embedded camera. 
To control some of the environmental conditions, we designed a template to use as the scene background (see Figure \ref{fig:plantilla}). 
During the acquisition process, however, we could not control some other environmental conditions, such as the capture viewpoint the camera setup or the illumination.

\begin{figure*}[!ht]
    \centering
    \includegraphics[height=140pt]{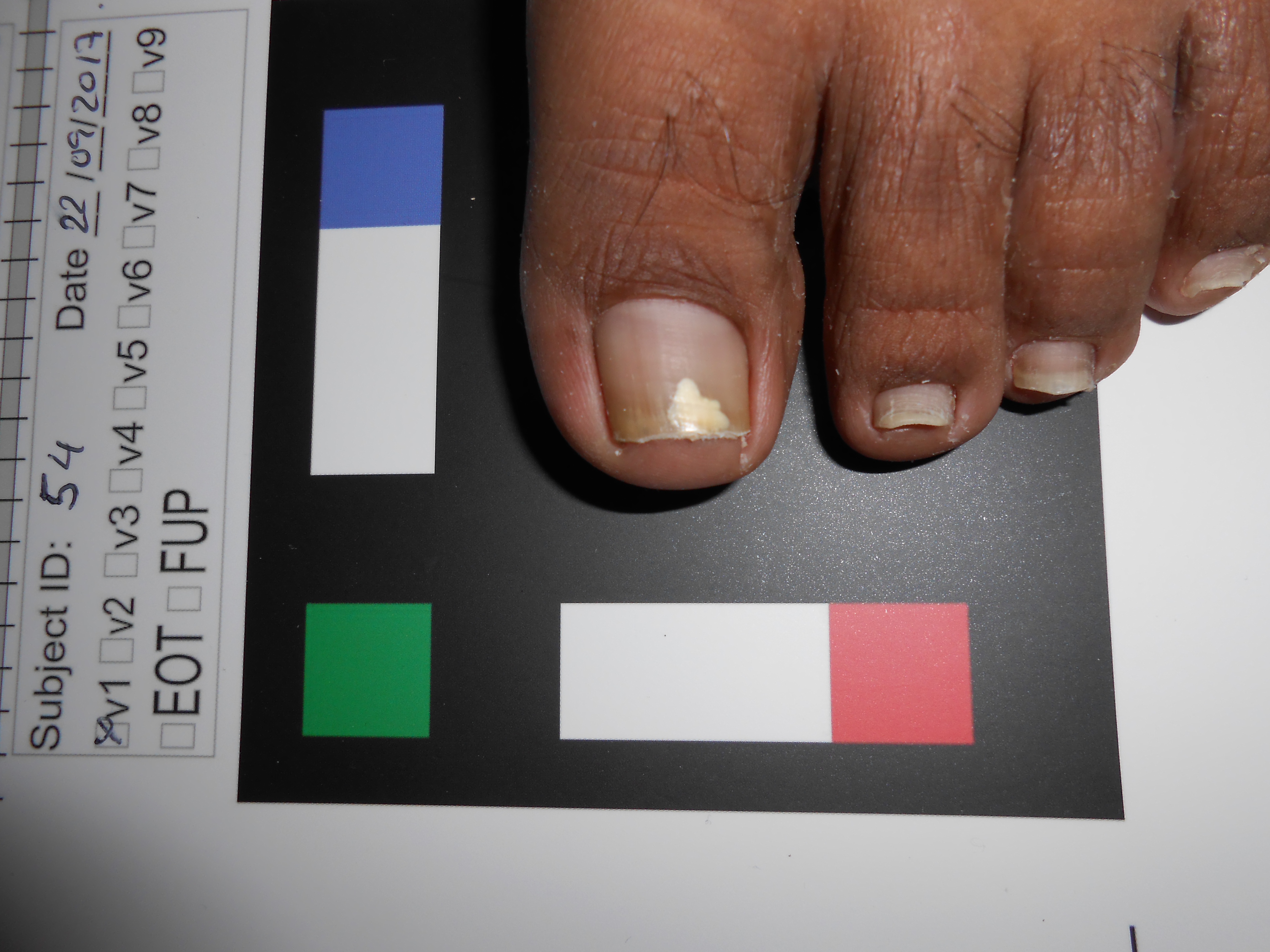}~
    \includegraphics[height=140pt]{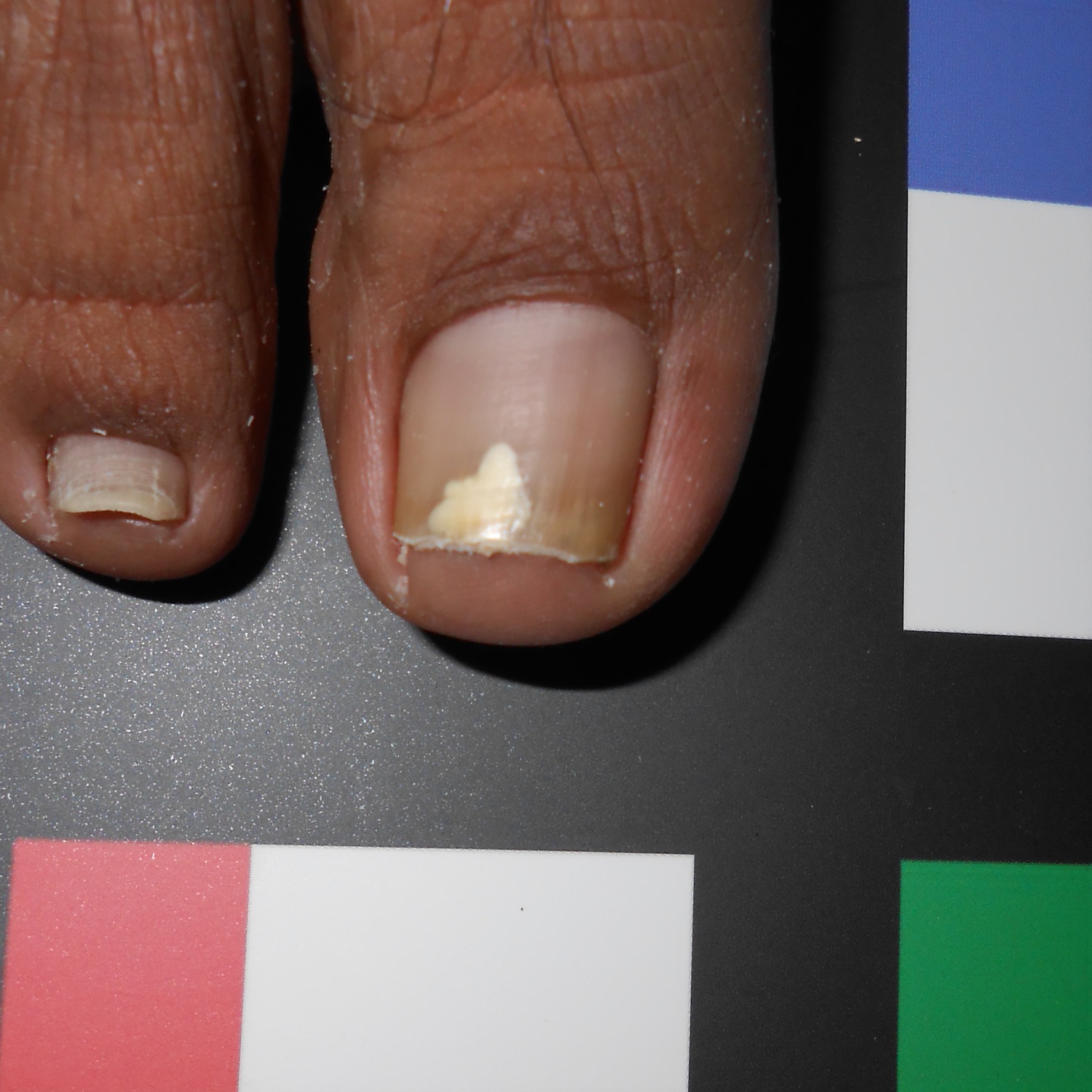}
    \caption{Original photograph of a left foot on the template (left) and its normalization (right).}
    \label{fig:normalization}
\end{figure*}

Before tackling the image segmentation problem, we perform an image normalization process based on the template's known measures. 
It consists on transforming the input image, as seen in Fig. \ref{fig:normalization} (left), to an image with standard dimensions and orientation, Fig. \ref{fig:normalization} (right). 
To achieve our objective we detect the position of the template corners and geometrically  transform the image with an affine mapping. 
As a result, all normalized images appear to had been taken under the same point of view.
We remark that the three template colored squares are mapped to the top-right, bottom-right and bottom-left image corners.
In particular, left foot images are mirrored.
Normalized images are always set to measure $1500\times 1500$ pixels.
Since the real region inside the template measures $5\times 5$ cm, a centimetre in the normalized image accounts for 300 pixels, which can be used to measure distances and areas.

\begin{figure*}[htpb!]
    \centering
    \includegraphics[width=0.7\textwidth]{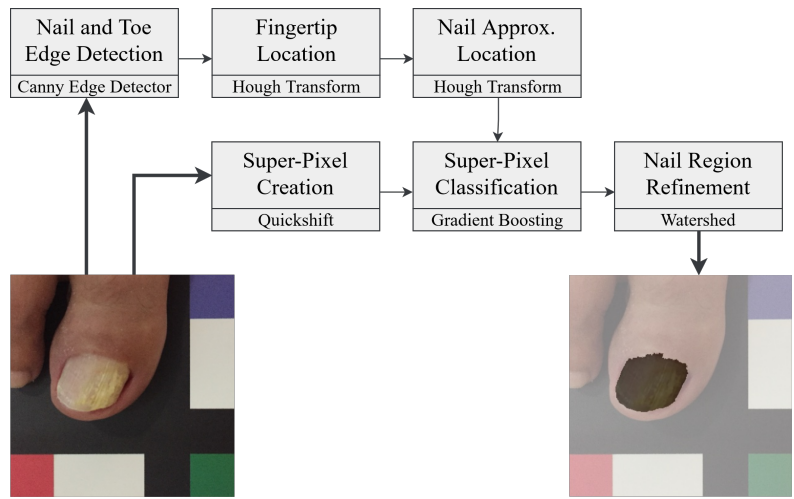}
    \caption{Flow Diagram of the robust toenail segmentation procedure.}
    \label{fig:FlowDiagram}
\end{figure*}

In the following, we describe each step taken to deal with the task of segmenting the toenail from images like the one in Fig. \ref{fig:normalization} (right).
In particular, Fig.\ \ref{fig:FlowDiagram} contains a flow diagram, along with the algorithm employed for each of them.

\subsection{Tip of the toe and nail circle estimation}
\label{sec:ToenailDetection:Circles}

In order to select the tip of the toe, we segment the foot regions from the background template.
Then, the \Houghtransform is used to detect a circle in the foot region, which corresponds to the tip of the toe.
Next, the nail is found within the tip of the toe using a second \Houghtransform, that depends on the result of the first one.
Specifically, this second \Houghtransform is computed taking into account only the edges that are in the tip of the toe circular estimation.
In the following, we detail the transformations performed. %, and we argument why they are needed.

First, we identify the foot Region of Interest (in short, foot ROI).
It is a mask that cover the part of the foot captured in the image, including both the skin and the nails.
Given the contrast with the black background template, this can be easily obtained based on pixel colorimetry: discriminating pixels based on a fixed range in the color space.
Specifically, we have used the skin color range given by Kovac \emph{et al.} \cite{kovac2003human}, and finally we select the biggest connected component as the foot ROI.
%This ROI is post-processed by removing all but the biggest connected component.

Next, we compute the edges on the foot ROI by employing several instances of the Canny edge detector \cite{DIP_gonzalez} using different values of the low threshold, $t_\text{low}$; the high threshold, $t_\text{high}$; and the standard deviation of the Gaussian filter, $\sigma$. We consider all the possible combinations of the following values.

\begin{equation*}%\label{eq:canny_params}
    \left\{
\begin{array}{rl}
    (t_\text{low},\ t_\text{high},\ \sigma)
    \:\:\big|\:\:
        & t_\text{low} \in \{ 5, 10, 15\},\\ 
        & t_\text{high} \in \{ 5, 10, 15, 20, 25\},\\ 
        & \sigma \in \{ 5, 10, 15\}\:
\end{array}
    \right\}
\end{equation*}

\begin{figure}[!ht]
\centering
        \begin{subfigure}[t]{0.2\textwidth}
                \centering
                \includegraphics[width=\textwidth]{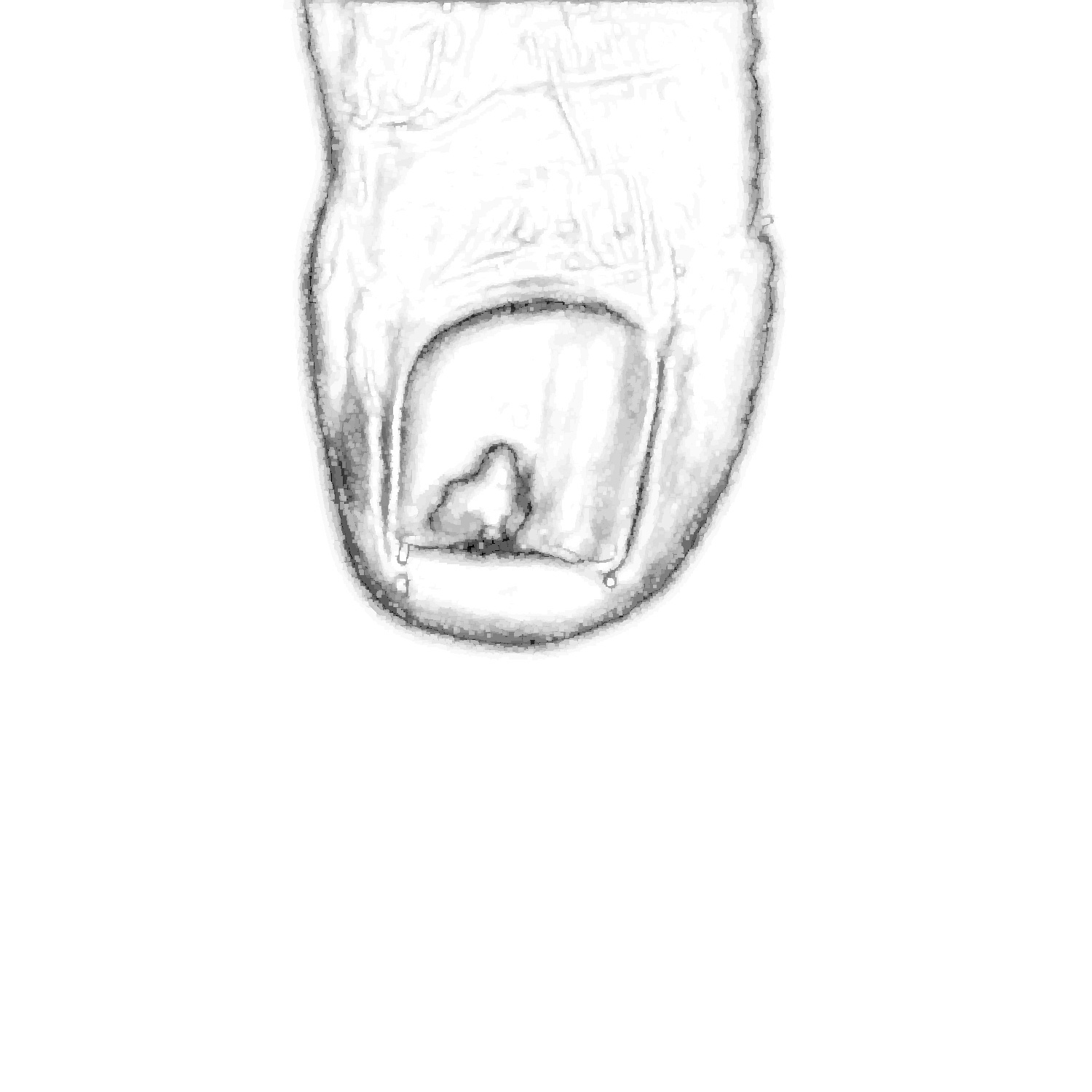}
                \caption{Foot edges.}
                \label{fig:canny_acum}
        \end{subfigure}
        % \hfill %add desired spacing between images, e. g. ~, \quad, \qquad etc.
        \begin{subfigure}[t]{0.2\textwidth}
                \centering
                \includegraphics[width=\textwidth]{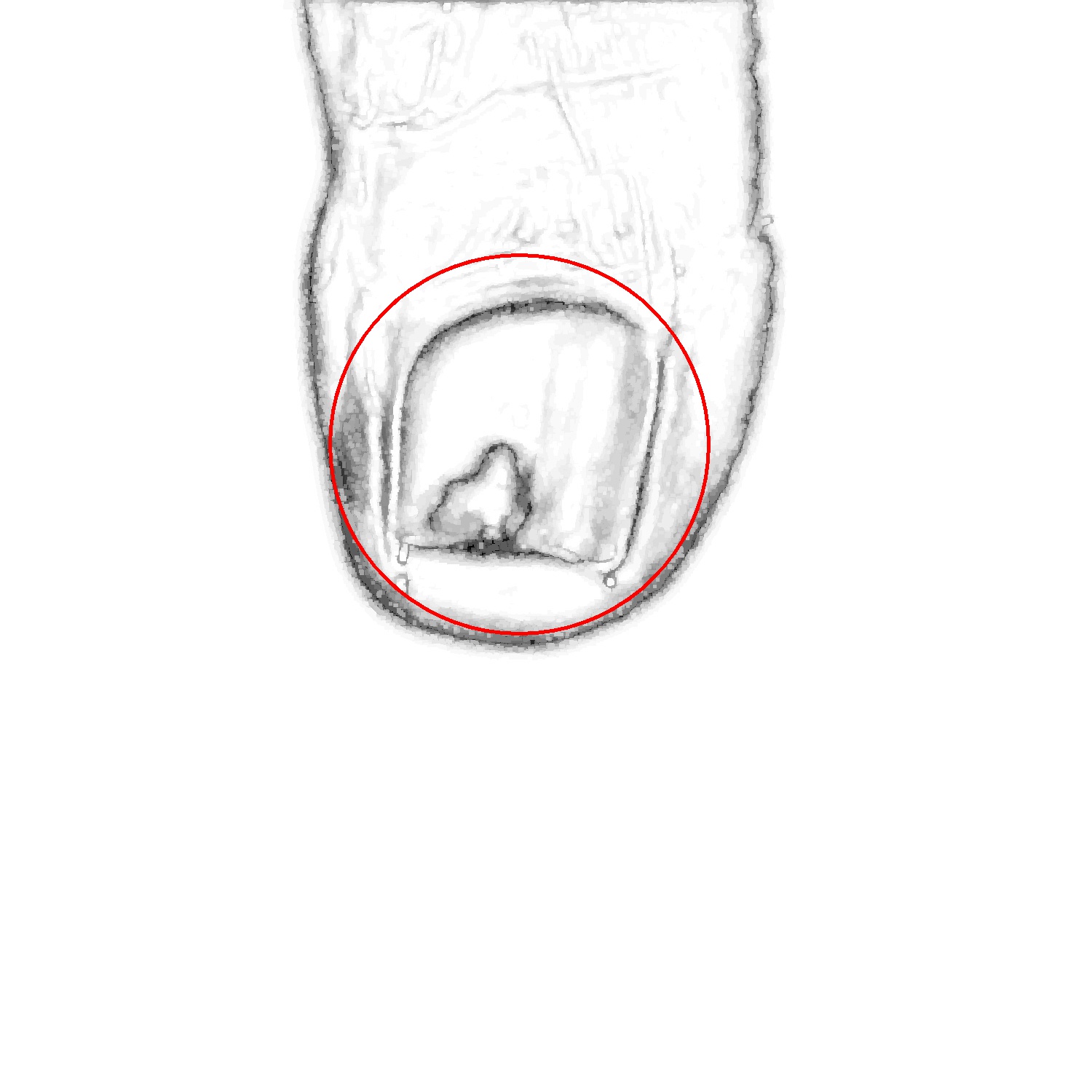}
                \caption{Tip of the toe best-circle estimation.}
                \label{fig:toe_circle}
        \end{subfigure}
        \\
        % \hfill %add desired spacing between images, e. g. ~, \quad, \qquad etc.
        \begin{subfigure}[t]{0.2\textwidth}
                \centering
                \includegraphics[width=\textwidth,clip]{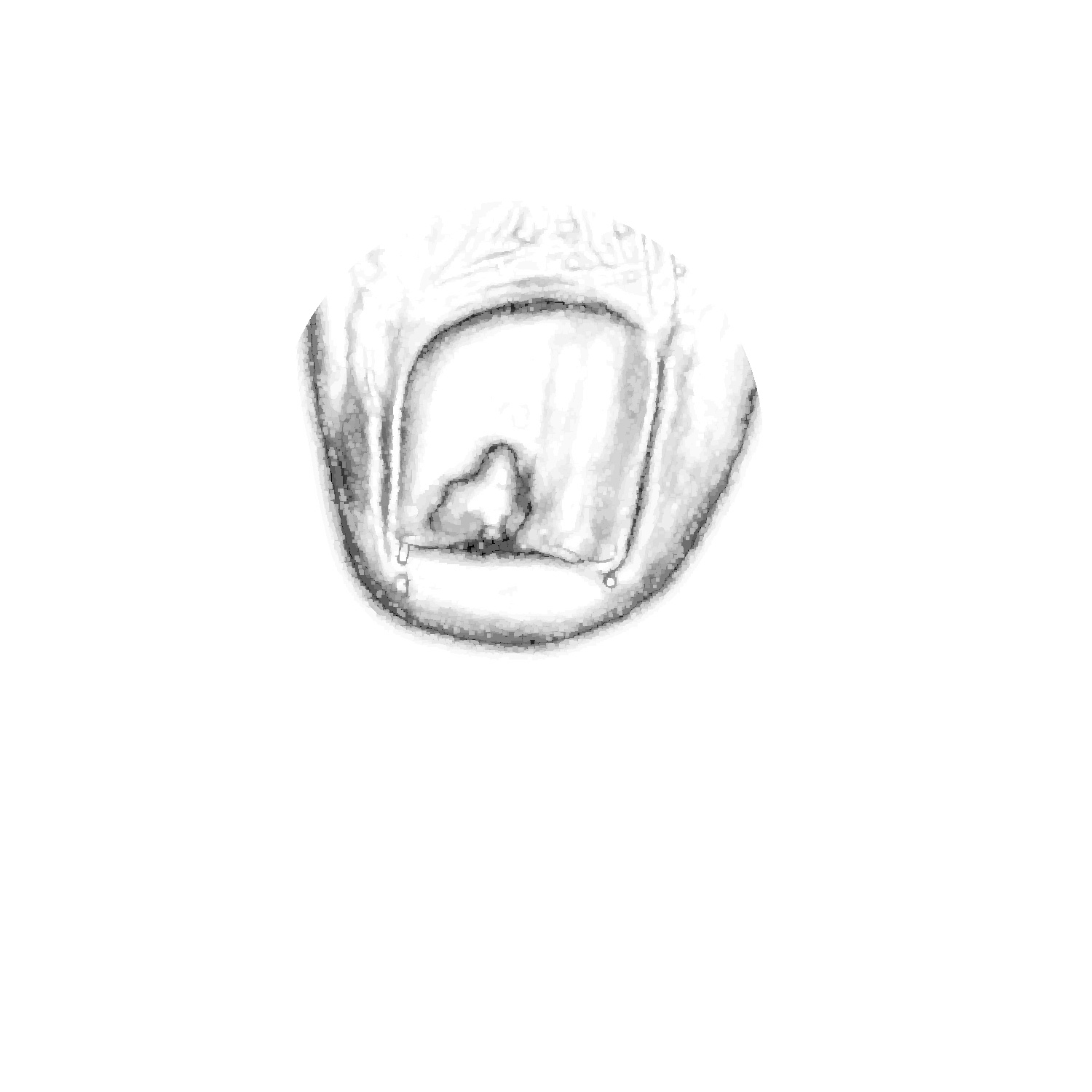}
                \caption{Nail edges.}
                \label{fig:canny_masked}
        \end{subfigure}
        % \hfill %add desired spacing between images, e. g. ~, \quad, \qquad etc.
        \begin{subfigure}[t]{0.2\textwidth}
                \centering
                \includegraphics[width=\textwidth,clip]{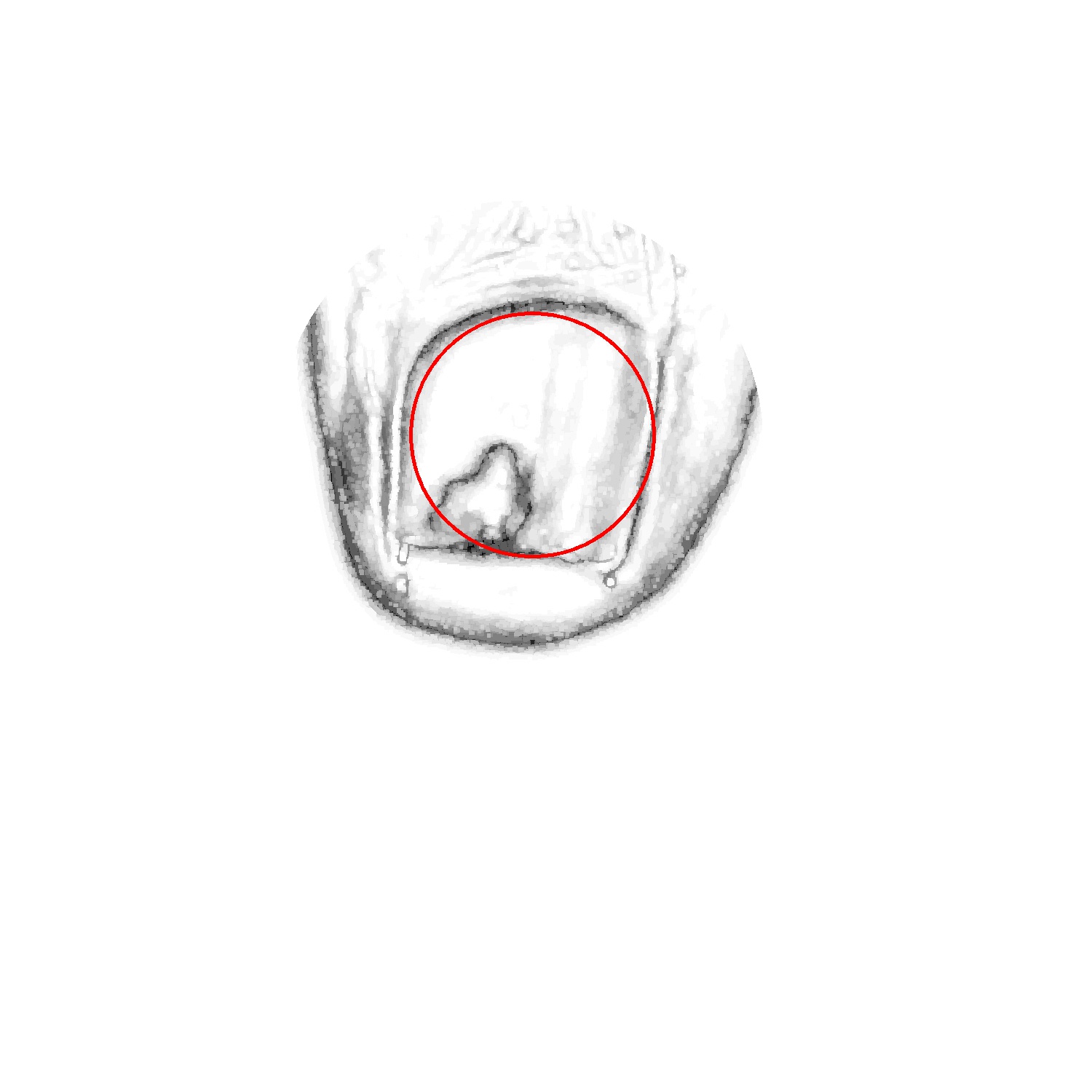}
                \caption{Nail best-circle estimation.}
                \label{fig:nail_circle}
        \end{subfigure}
    \caption{Edges and figures computed in tip of the toe detection process.}
    \label{fig:circles}
\end{figure}

By adding up each one of the Canny edge instances we obtain a cumulative contour image, shown in Fig. \ref{fig:canny_acum}.
%This has experimentally proven to be a robust approach to detect all edges in the image, including those of the nail.

Onto that edge image, we apply the Hough transform to locate the tip of the toe.
Due to its position and size, the area inside this pattern contains the nail.
So, we apply the circular \Houghtransform on the edge image and we select the best candidate as the nearest circle to the template bottom right corner that has a predefined percentage of its area within the foot ROI (specifically, we choose a threshold of $0.85$). Also, the computed circles are limited to radius between 0.85 cm and 1.75 cm.
The selected circle is shown in Fig. \ref{fig:toe_circle}. 
We remark that this circular pattern captures the tip of the toe but does not locate the nail with acceptable accuracy.

Finally, we detect the circle that better fits the nail using a second \Houghtransform.
We discard the edges far from the tip of the toe circle (see Fig.\ \ref{fig:canny_masked}), so that we mainly keep the nail edges.
Thus, this second \Houghtransform is prone to detect the circle that best fits the nail.
Also, we constrain the radius of this circle according to the size of the tip of the toe.
Specifically, we expect the nail radius to be smaller than the radius of the tip of the toe circle (see Fig.\ \ref{fig:toe_circle}) but bigger than half of its measure. 
The most prominent circle (see Fig.\ \ref{fig:nail_circle}) is the one selected as nail circle.
Experimentally, the results of this process have proved to always find a location on the nail (the circle center) and a good estimation of the nail size (derived from the radius).

\subsection{Super-Pixel Classification}
\label{sec:ToenailDetection:Superpixel}

The process explained above does not provide enough information to accurately segment the nail.
A good approach for refining these results is to use  machine learning methods. 
%A first attempt based on the color of individual pixels proved to be insufficient due to the similarity of the nail and skin textures, as previously mentioned.
To overcome this, we identified groups of pixels with some common characteristics.
Considering these groups as entities to classify, our problem is simplified but enough information is kept to ensure it is still attainable.

\begin{figure*}[!ht]
\centering
        \begin{subfigure}[t]{0.3\textwidth}
                \centering
                \includegraphics[width=\textwidth]{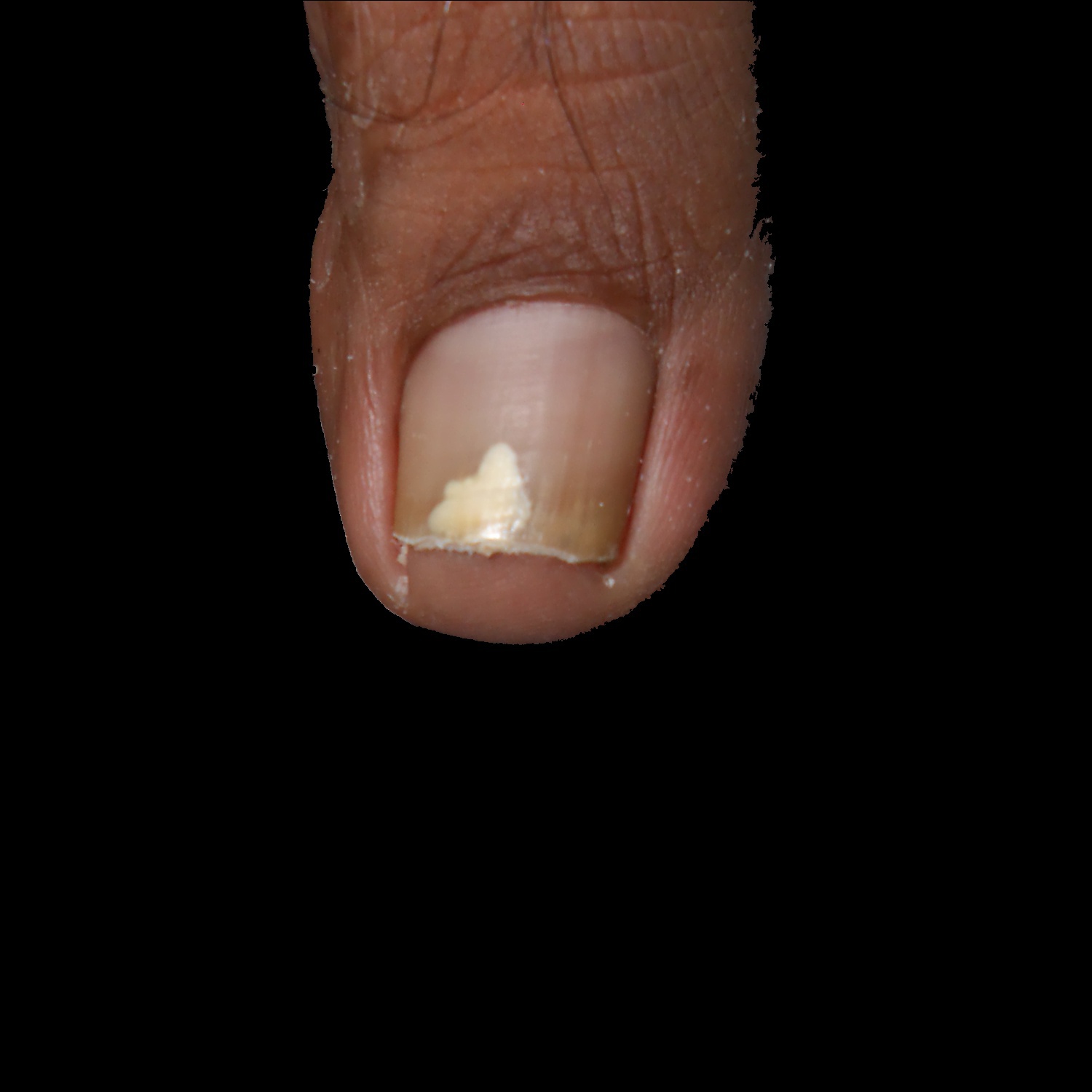}
                \caption{Normalized image with the foot ROI masked out.}
                \label{fig:img_only_finger}
        \end{subfigure}
        ~ %add desired spacing between images, e. g. ~, \quad, \qquad etc.
        \begin{subfigure}[t]{0.3\textwidth}
                \centering
                \includegraphics[width=\textwidth]{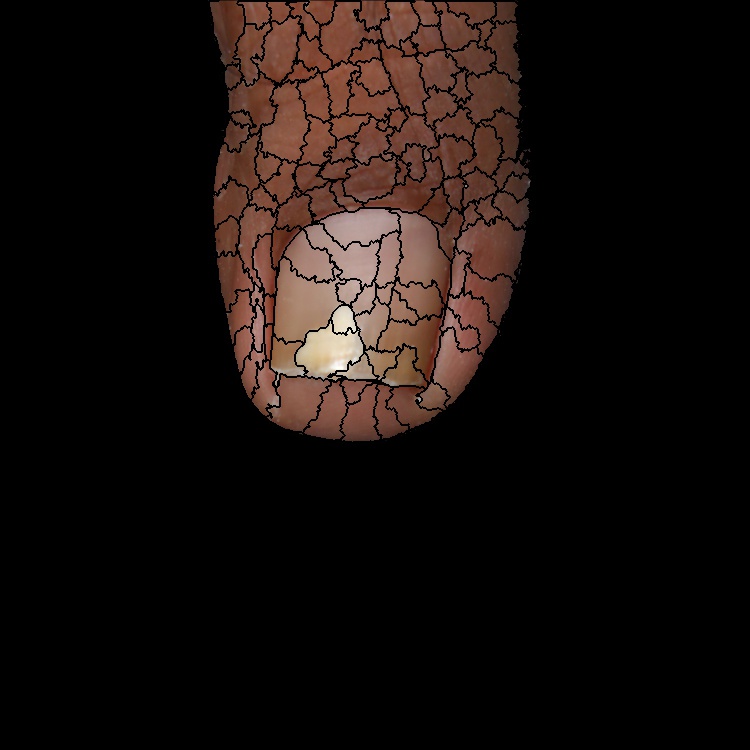}
                \caption{Quickshift output of the image in Fig. \ref{fig:img_only_finger}.}
                \label{fig:quickshift_img}
        \end{subfigure}
        ~ %add desired spacing between images, e. g. ~, \quad, \qquad etc.
        \begin{subfigure}[t]{0.3\textwidth}
                \centering
                \includegraphics[width=\textwidth,clip]{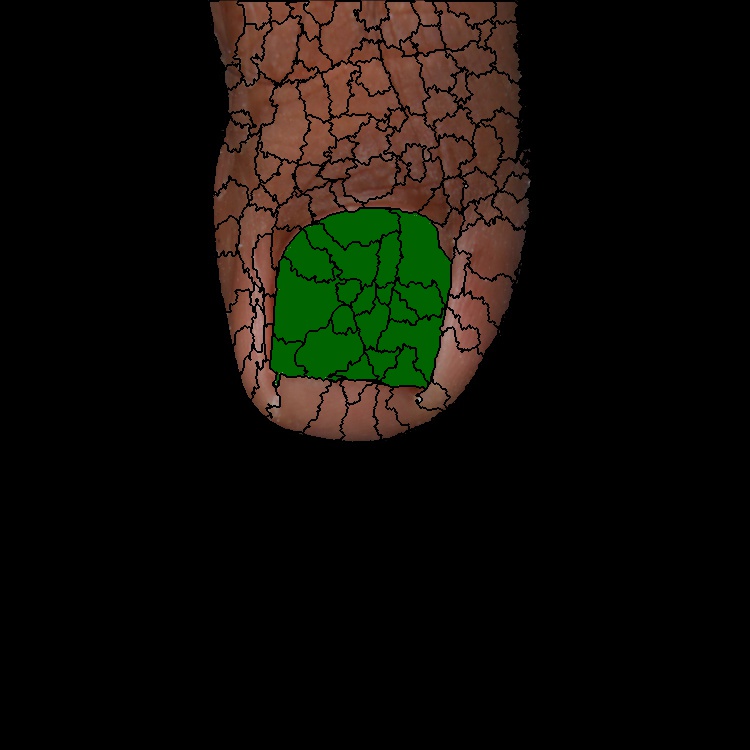}
                \caption{Nail super-pixels tagged on Fig. \ref{fig:quickshift_img}.}
                \label{fig:clf_img_res}
        \end{subfigure}
\caption{Super-pixel approach to identify regions as nail or toe.}
\label{fig:Samples}
\end{figure*}

To group close and similar pixels we use the Quickshift algorithm \cite{vedaldi2008quick}.
It divides the image in connected and uniform regions, the so-called super-pixels.
%It is out of the scope of this paper to describe the Quickshift algorithm in detail, and we refer the reader to the work by Vedaldi and Soatto \cite{vedaldi2008quick}.
However, as can be seen in Fig.\ \ref{fig:Samples}, a set of connected super-pixels define the nail contour accurately.

%nail contours are typically well defined and, thus, always separate super-pixels.
%In other words, 

\subsubsection{Classifier Features}

In this section, we introduce the features used to classify if a super-pixel is part of a nail or not.
These features are of a different nature: some come from the tip of the toe detection step, leveraging the best-circle nail estimation and the foot ROI; others from the colorimetry of pixels; and we also consider other geometric attributes of the super-pixels as a region.
% The objective of the trained model is split super-pixels between nail ones and skin ones. 
For the sake of completeness  Table~\ref{table:summary_clf_features} list all used features.

% \usepackage{booktabs}
% \usepackage{multirow}
\begin{table*}[!ht]
\centering
    \begin{tabular}{@{}llc@{}}
        \toprule
        \multirow{2}{0.10\textwidth}{\parbox{0.12\textwidth}{
        \centering Features based on
        }} & \multicolumn{1}{c}{Description} & 
        \multirow{2}{0.10\textwidth}{\parbox{0.10\textwidth}{
        \centering Number of features
        }} \\ 
        ~ \\
        \midrule
        \multirow{2}{*}{Colorimetry}      & 
        \begin{tabular}[c]{@{}l@{}}
            Mean of super-pixel pixels channels in \texttt{HSV}, \texttt{RGB} and and \texttt{CIELAB} colorspaces.
        \end{tabular}                                                    
        & 9 \\ 
        \addlinespace
        & \begin{tabular}[c]{@{}l@{}}
            Standard deviation of super-pixel pixels channels in \texttt{HSV}, \texttt{RGB} and \\
            \texttt{CIELAB}.
            \end{tabular} 
        & 9 \\ 
        \addlinespace
        \hline
        \addlinespace
        Region information  & Super-pixel area and perimeter. & 2 \\ 
        \addlinespace
        \hline
        \addlinespace
        \multirow{4}{*}{Position}       & Super-pixel centroid. & 2 \\
        \addlinespace
        & \begin{tabular}[c]{@{}l@{}}
            Distance between super-pixel centroid and the rightest and \\
            bottommost skin pixel in the same centroid altitude.
            \end{tabular}
        & 2 \\ 
        \addlinespace
        & Distance to circles.    & 6 \\ 
        \addlinespace
        \hline
        \addlinespace
        Image circles data  & Radius of two circles. & 2 \\ 
        \addlinespace
        \bottomrule
    \end{tabular}
\caption{Classifier features.}
\label{table:summary_clf_features}
\end{table*}

%\subsubsection*{Colorimetry}

We use channels of different color spaces as features. 
Visually, we can distinguish nail from the rest of the skin, and so we hypothesize that the colorimetry may provide some information about a super-pixel's relevance.
We use the following color spaces: \RGB, \HSV and \CIELab.

% \begin{description}
%     \item[\RGB.] 
%     This color space represents the contribution of red, green and blue components of the color. 
%     It has a high correlation between the channels, which is a disadvantage if we are using its components to define features.
%     It is not perceptually uniform.

%     \item[\HSV.] 
%     It represents the hue, saturation and value channels.
%     They define, respectively, the perceived color of an area, the purity of the color and its brightness.

%     \item[\CIELab.] 
%     This color space was proposed by the International Commission on Illumination (CIE, in French), whose main goal was to provide a uniform color space. 
%     This means that the Euclidean distance between two \CIELab colors is strongly correlated with the human visual perception.
%     Another advantage of this color space is the separation of the luminance component, \channelname{L}; from the chromatic channels, \channelname{a}, \channelname{b}. 
% \end{description}

The size of the area occupied by the super-pixel region and its perimeter may also be useful. 
Also, we foresee that there may exist a correlation between these characteristics and the rest of them.
For instance, their size may provide an indicator of how variable the colorimetry is in such a region.

%\subsubsection*{Position}

We also consider features related with the centroid's position of the super-pixel. 
Concretely, we consider the position in the $X$ and $Y$ axis, the distance to the rightest skin pixel in the same row and the distance to the bottommost skin pixel in the same column, and the distances to both circle centers---the toe tip circle and the estimated nail circle.

We remark that the radius of these two circles are also included as features.
%They hold the exact same value for many super-pixels, but it varies when considering a different image.
% We have the distance to the circle centroid in X axis and in Y axis.

\subsubsection{Classification}

% Now, that we have defined a set of features, we are ready to apply machine learning. We prepare our data and split it in a train and test set, built from different datasets data. All the training process are made with the training set, and finally tested with the test set.
Different classifiers have been used with the previously mentioned features to classify the super-pixels.
We have experimented with the following classifiers: \SVM, \RandomForest, \GradientBoosting and
\MultilayerPerceptron.

% \begin{description}
%     \item[\SVM.] Statistical learning classifier that is well-formalized and widely used.
%     It builds a hyper-plane that separates multidimensional data according to the defined classes \cite{burges1998tutorial}.
    
%     \item[\RandomForest.] Ensemble of decision trees classifiers.
%     It fits a potentially large number of decision trees on a series of sub-samples of the dataset.
%     It averages their result to improve the predictive accuracy and control over-fitting \cite{randomForest}.
    
%     \item[\GradientBoosting.] Ensemble of classifiers that builds an additive model in a forward stage-wise fashion \cite{GradientBoosting}.
%     In each stage, it fits a number of regression trees using the negative gradient of the deviance loss function. 
%     The binary \GradientBoosting classifier is a particular case where only a single regression tree is trained \cite{GradientBoosting}. 
%     It is able to optimize arbitrary differentiably loss functions. 
    
%     \item[\MultilayerPerceptron.] One of the most commonly used architectures of artificial neural network to learn from hard-coded features. 
%     They are parallel distributed systems composed of layers of simple and independent processing units linked by weighted connections \cite{haykin1994neural}.
% \end{description}

%The results obtained are described in Table \ref{tab:results_clfs}. 

\subsubsection{Comparison of the classifiers}
\label{sec:ToenailDetection:ComparisonClassifiers}

Table \ref{tab:results_clfs} summarize the results obtained with different classifiers, we used the literature recommended parameters for each classifier. The results are the average performance measure across test samples. %(see Sect.\ \ref{sec:Results:Settings}).
%by their original authors or 
%without fine-tuning its hyperparameters but instead balancing their execution time.
The performance metrics are computed using the test set, previously unseen during the training stage.

\begin{table*}[!ht]
    \centering
    \caption{Performance metrics obtained with different classifiers, where the best value is highlighted for each metric.}
    \label{tab:results_clfs}
    \begin{tabular}{lcccccc}
    \toprule
        Classifier &  \Sensitivity &  \Specificity &  \Precision &  \Accuracy &     \Fmeasure &  \CohensKappa \\
    \midrule
        \RandomForest &        0.983 &        0.896 &      0.888 &     0.971 &  0.892 &        0.875 \\
        \MultilayerPerceptron &        \textbf{0.984} &        0.898 &      \textbf{0.893} &     0.972 &  0.896 &        0.880 \\
        \GradientBoosting &       0.983 &  0.921 &     0.891 &     \textbf{0.975} &  \textbf{0.906} &        \textbf{0.891} \\
        \SVM &        0.964 &        \textbf{0.958} &      0.804 &     0.964 &  0.874 &        0.853 \\
    \bottomrule
    \end{tabular}
\end{table*}

According to the results in Table \ref{tab:results_clfs}, Gradient Boosting appears to be the best classifier.
Let us emphasize the fact that each performance measure is affected differently by deviations to the ground truth. %(see Sect.\ \ref{sec:Results:Settings} for a detailed description of the performance measures).
In particular, \Sensitivity\ and \Precision\ are only affected by false positives, and \Specificity\ by false negatives.
On the other hand, the rest of measures provide a more balanced insight of the overall performance of the method.
We specifically rely on the \Fmeasure\ and the \CohensKappa\ due to their robustness: they successfully handle both types of errors and class imbalance problems.
Thus, \GradientBoosting is used as our default training model during the rest of the paper. 
We observe that the \MultilayerPerceptron, after an exhaustive hyper-parameter fine-tuning, could become a good alternative.
However, we favour Gradient Boosting as the default training model due to its ease of deployment.
% \todo{La seguent frase crec que ja s'ha dit a paragrafs anteriors}
% We observe that, for the sake of obtaining a fair comparison, the four classifiers are trained and tested without fine tuning their hyperparameters.

\subsection{Toenail segmentation}
\label{sec:ToenailDetection:Watershed}

A final step is presented here to refine the nail location using the \watershed algorithm \cite{meyer1992color}.
%To sum up, we have already detected the tip of the toe of the big toe with the Hough transform, the circle that best fits the nail with the same technique, and we have classified super-pixel regions with the \GradientBoosting algorithm.
%Each step is based on the previous one, and provides higher quality information.
%In this final stage, we use the watershed transform to further improve the results.
Implicitly, we leverage the fact that nails are very well defined by their frontier, much better than by their colorimetry, size or shape.

The \watershed segmentation algorithm requires the definition of initial markers that grow until they fill a whole region.
To initialize the algorithm, we use the probabilities of each super-pixel to belong to a class, given by the \GradientBoosting classifier.
More specifically, we initialize some super-pixels as \watershed initial markers as follows:

\begin{itemize}
    \item Marked as \emph{background}. Super-pixels on the  excluded region of the foot ROI mask, slightly eroded with $5x5$ kernel. %%to avoid marking the frontier.
    \item Marked as \emph{nail}. Super-pixels whose estimated probability of being part of the nail is greater than 99.99\%, to guarantee as much as possible the correctness of the initial marker.
    \item Marked as \emph{skin}. Super-pixels with a probability greater than 99.99\% of being skin.
    \item \textit{Unmarked}. The rest of pixels are tagged by the \watershed algorithm based on their closeness and similarity to the already marked pixels.
\end{itemize}

\begin{figure*}[!ht]
\centering
    \begin{subfigure}[t]{0.21\textwidth}
            \centering
            \includegraphics[width=\textwidth]{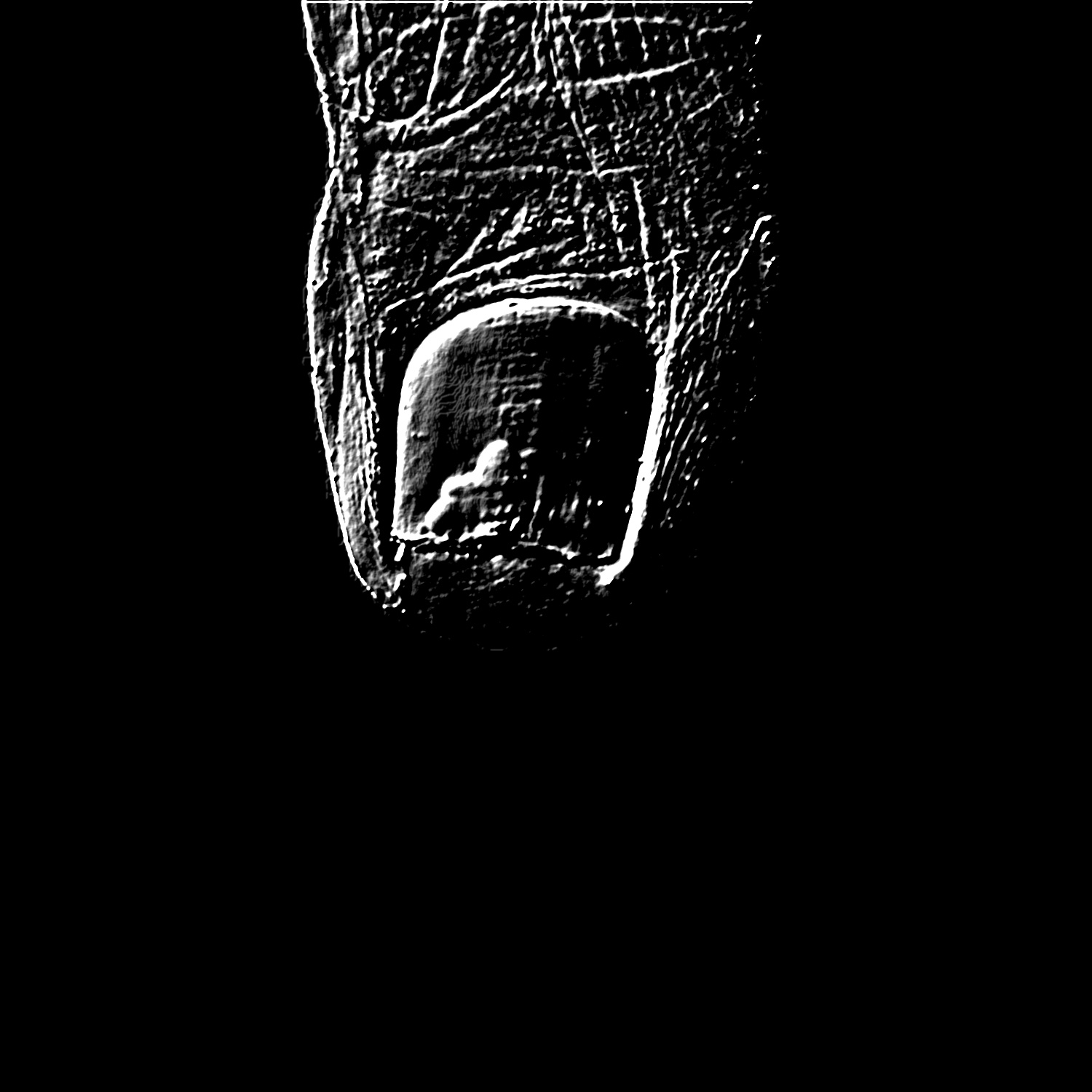}
            \caption{Gradient image.}
            \label{fig:gradient}
    \end{subfigure}
    ~
    \begin{subfigure}[t]{0.21\textwidth}
            \centering
            \includegraphics[width=\textwidth]{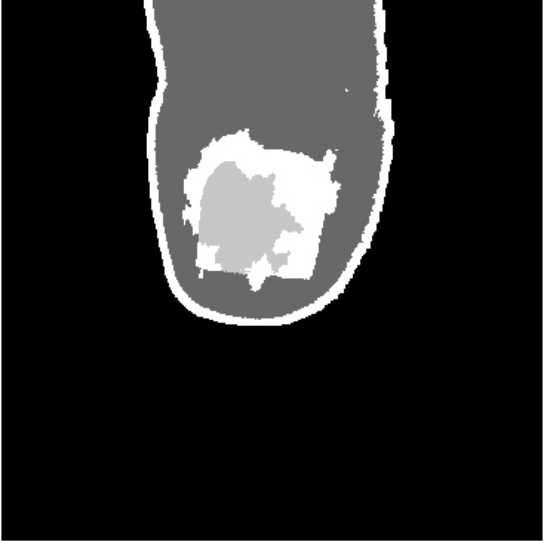}
            \caption{Initial markers, with white representing unmarked regions.}
            \label{fig:initial_markers}
    \end{subfigure}
    ~
    \begin{subfigure}[t]{0.21\textwidth} 
            \centering
            \includegraphics[width=\textwidth]{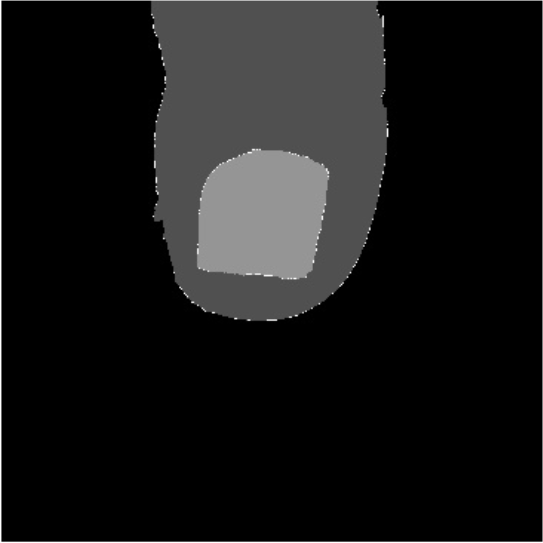}
            \caption{Final markers.}
            \label{fig:final_markers}
    \end{subfigure}
    ~
    \begin{subfigure}[t]{0.21\textwidth}
            \centering
            \includegraphics[width=\textwidth,clip]{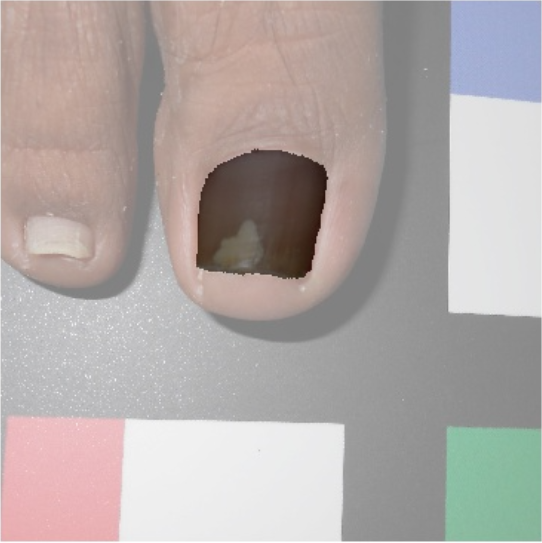}
            \caption{Nail mask from Fig.\ \ref{fig:final_markers} overimposed on the original image.}
            \label{fig:result_mask}
    \end{subfigure}
    \caption{Watershed-based nail detection flow.}
    \label{fig:watershed}
\end{figure*}

%The \watershed algorithm fills the unmarked regions until it reaches a high gradient that works, metaphorically, as a barrier. 
In Fig. \ref{fig:gradient}, we can appreciate that the contours of the nail are sharp. 
Thus, provided the initial markers are corrected, the growth of the nail and skin regions would be prone to stop at these edges.
This has been actually the case when processing images in practice, as shown in Fig. \ref{fig:result_mask}.

\section{Experimentation and Results}
\label{sec:Results}

In this section, we study the results provided by the algorithm and, when appropriate, discuss the design decisions taken during its development.
We analyze which are the more informative features to distinguish between nail and toe regions, which proved to be the hardest problem in practice.
Also, we introduce quantitative performance measures to provide objective indicators of \emph{how successful} the proposed method is.

\subsection{Experimental framework}
\label{sec:Results:Settings}

\begin{figure*}[!ht]
\centering
    \begin{subfigure}[t]{0.31\textwidth}
            \centering
            \includegraphics[height=120pt]{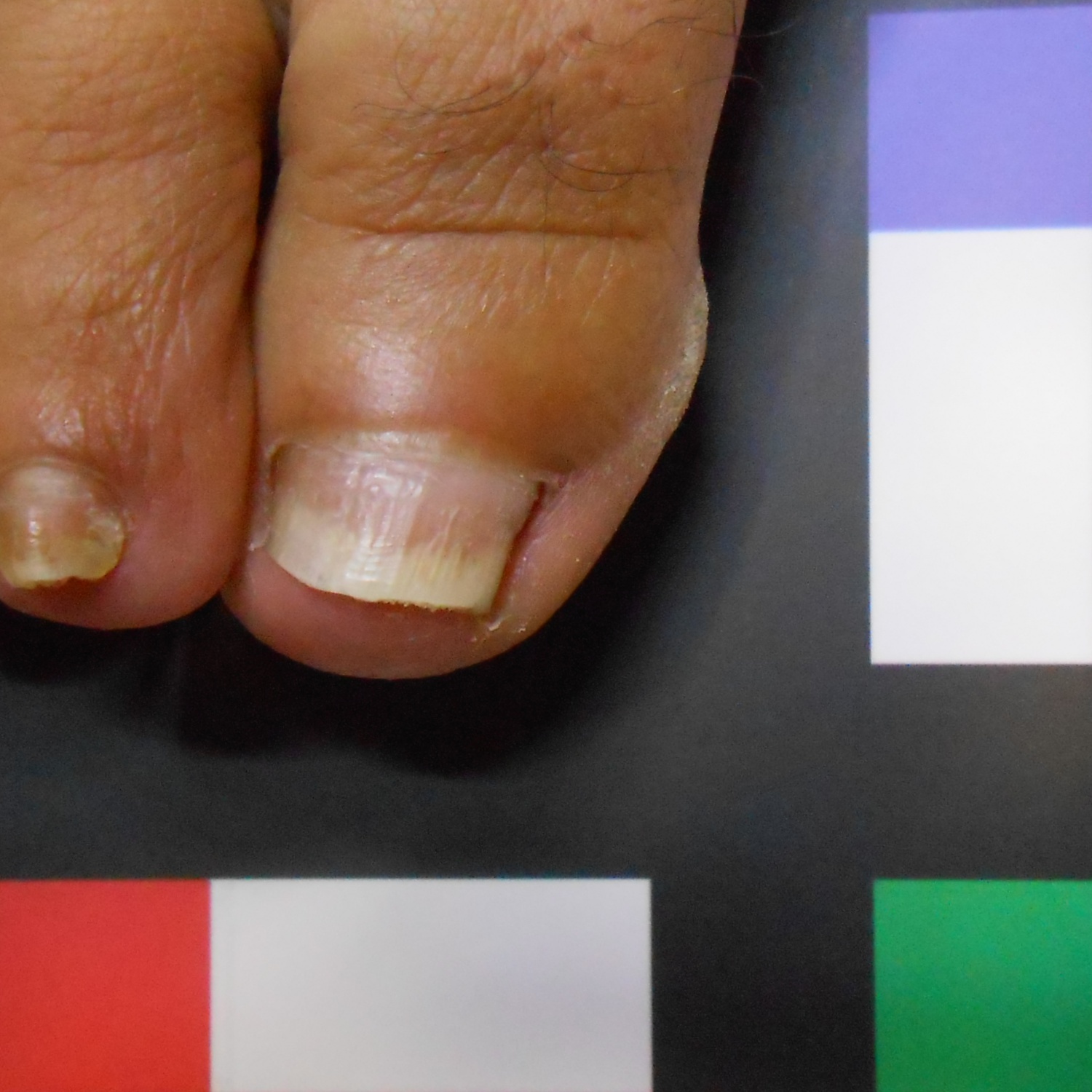}
            % \caption{Input image.}
            % \label{fig:original}
    \end{subfigure}
    ~
    \begin{subfigure}[t]{0.31\textwidth}
            \centering
            \includegraphics[height=120pt]{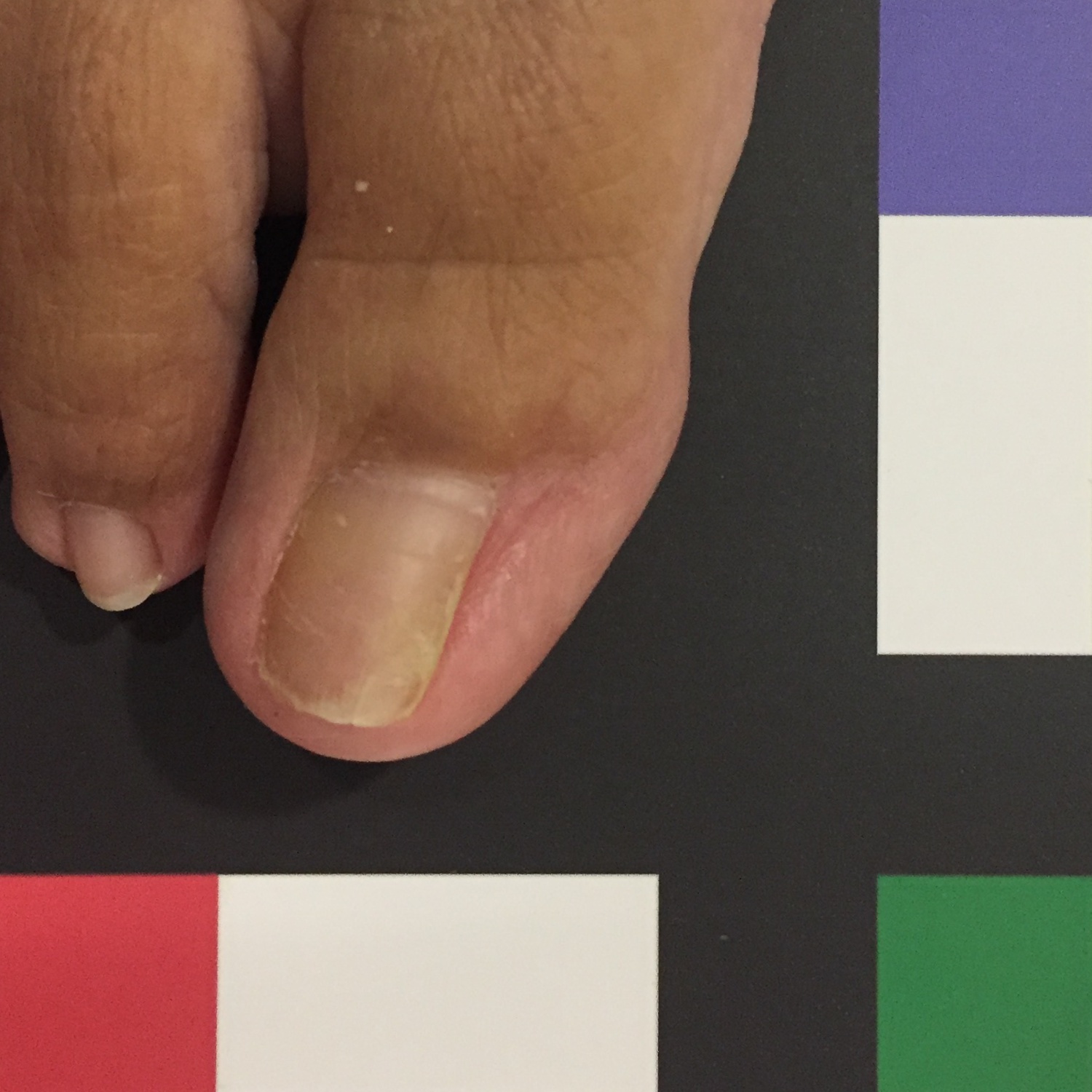}
            % \caption{Input image.}
            % \label{fig:original}
    \end{subfigure}
    ~ %add desired spacing between images, e. g. ~, \quad, \qquad etc.
    \begin{subfigure}[t]{0.31\textwidth}
            \centering
            \includegraphics[height=120pt]{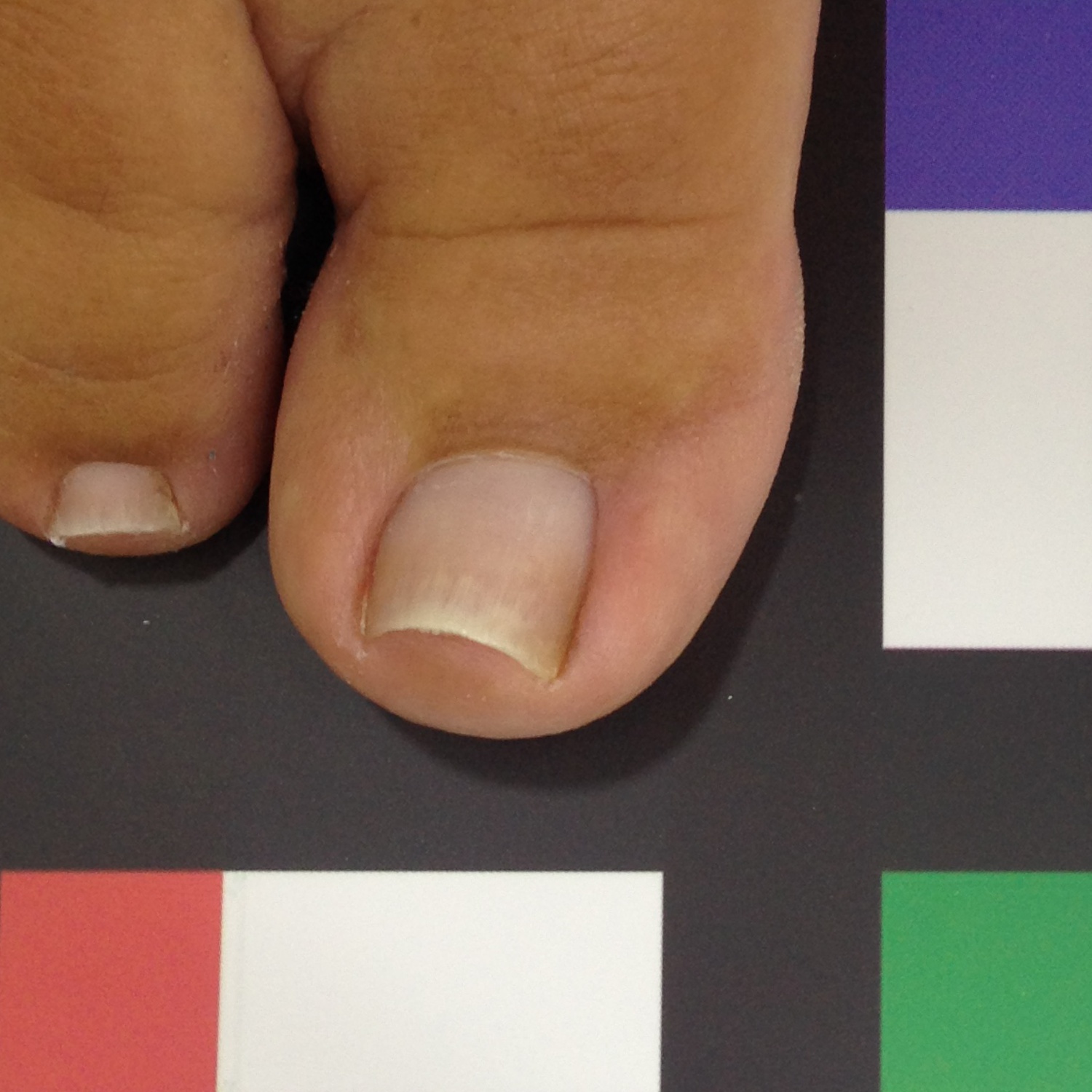}
            % \caption{Input image.}
            % \label{fig:original}
    \end{subfigure}
\caption{Dataset examples after the normalization process.}
\label{fig:examples}
\end{figure*}

The dataset is composed of 348 images of human big toes acquired using the doctor's cameras attached to off-the-shelf smartphones.
A sample image is found in Fig. \ref{fig:examples}.
As previously explained, during the image acquisition stage, some parameters could not be controlled, such as the illumination, the specific capture viewpoint and the camera specifications.

To accurate evaluate the method's performance, we divide the images in two sets: 257 images for training and 91 for testing. %(roughly 25\%).
%The former one is used to design and train the algorithms, whereas the latter is reserved to obtain a non-biased estimator of the performance of the method.
All of them have a manually segmented ground truth which separates the classes of toe, nail and surrounding background (corresponding to the template employed).

Several performance metrics are used to quantify the results obtained.
They complement the visual representations, such as the one in Fig. \ref{fig:watershed}, since they exhibit the results over the whole dataset. 
%We emphasize that a single estimator is not useful for every application, since they are affected by different types of error.
Since we are dealing with a segmentation task, we employ pixel-wise measures.
The performance measures used are sensitivity, specifity, accuracy, precision, F-measure and Cohen's kappa. They complement each other: each of them captures distinctly different deviations from the expected result.

% which are based on dividing the pixels in four categories: true positives (\TP), nail pixels correctly tagged; true negatives (\TN), toe pixels correctly tagged; false positives (\FP), toe pixels tagged as nail; and false negatives (\FN), nail pixels wrongly labelled as part of a toe.
%We emphasize that the surrounding black background, corresponding to the template (see Fig. \ref{fig:examples}), is easily discarded and, thus, not taken into account to compute the metrics.

% The performance measures used are: 
% the \emph{\Sensitivity} (or \emph{True Positive Rate}), that measures the proportion of positives correctly identified; 
% the \emph{\Specificity} (or \emph{True Negative Rate}), that measures the proportion of negatives correctly identified; 
% the \emph{\Accuracy}, that measures the proportion of correctly identified samples;
% the \emph{\Precision}, that measures the proportion of positives samples identified as such;
% the \emph{\Fmeasure}, which is the harmonic mean of \Accuracy\ and \Precision;
% the \emph{\CohensKappa}, a statistic that measures the agreement between classes.

% We include a wide range of metrics since all of them capture different types of errors, none of them being better than the others in an absolute sense. 
% For instance, we could obtain a perfect \Sensitivity\ score with the naive method of estimating all pixels as positive.
% By considering a set of them, we provide a good, objective measurement of the overall performance.

\subsection{Method's performance}
\label{sec:Results:Analysis}

In this section we expose and explain the results obtained after running the classification stage of the method.
%We briefly recapitulate some of the stages of our method (see also Fig.\ \ref{fig:FlowDiagram}).
%First, we split the image in super-pixels using the \quickshift algorithm. 
%Second, we classify the super-pixels using the \GradientBoosting classifier. 
%Third, we select the super-pixels with most probability of belonging to a class, and we use them as initial markers for the \watershed algorithm. 
%Finally, we set the nail region to be the output of the \watershed algorithm.
% \todo{En la següent subsecció, els resultats es donen executant TOT el mètode (incloent el watershed), o es donen els resultats només del First step -> Dues següents subseccions només classificadors, la quantitative performance és la completa.}

\begin{figure}[!ht]
    \centering
    \includegraphics[width=0.5\textwidth]{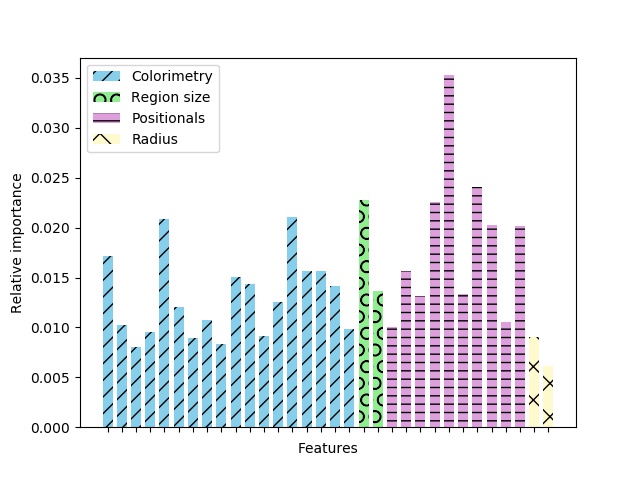}
    \caption{Relative feature importance according to the \GradientBoosting classifier.}
    \label{fig:bars_importances}
\end{figure}

%\subsection{Feature importance}
To study the task at hand, we analyse the importance of each feature used to distinguish nail and toe regions.
In Fig. \ref{fig:bars_importances}, we can see the importance of the features in the \GradientBoosting classifier. 
Such feature importance is averaged over the trees that conform the \GradientBoosting ensemble.
In individual trees, it is a measure of how well each variable splits the data at a specific node \cite{hastie2009elements}.

%We observe that the colorimetry represents the larger group of features used by the training models. 
%The two most informative ones in this group are the mean and the standard deviation of the \channelname{a} channel from the \CIELab color space.
%In other words, this appears to be the more informative channel toward separating the two classes.
%The mean of Hue channel in the \HSV color space also provides a significant amount of information.

We observe that the most informative feature of the classifier is, however, a positional feature: the distance between a super-pixel centroid and the center of the estimated nail circle.
Again in Fig. \ref{fig:bars_importances}, we observe that the difference between this feature importance and the other ones is enormous in relative terms.
This is because the regions identified as nail are the ones closer to the circle, and the distant ones can be straightforwardly discarded.

The colorimetry represents the larger group of features used by the training models. The two most informative ones in this group are the mean and the standard deviation of the \channelname{a} channel from the \CIELab color space.

Other features that give important information are the super-pixel area, and the distances in the \channelname{Y} axis to the nail circle and to the lowest skin pixel.

\begin{figure}[!ht]
    \centering
    \includegraphics[width=.5\textwidth]{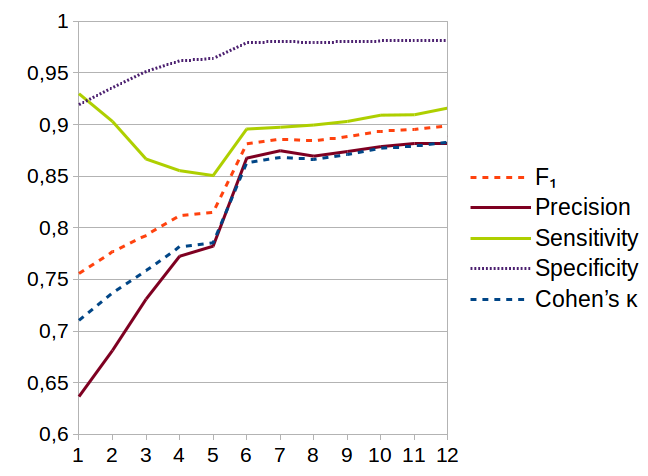}
    \caption{Different performance metrics for the \GradientBoosting classifier with a restricted number of features, ordered by their relative importance.}
    \label{fig:metrics_evolution}
\end{figure}

In Fig. \ref{fig:metrics_evolution}, we can appreciate the evolution of the method accuracy when sequentially adding features to the \GradientBoosting classifier. 
The features are added in the order of importance already shown in Fig. \ref{fig:bars_importances}.

Table \ref{tab:results_rates} presents the results obtained by evaluating the final mask---provided by the watershed algorithm---against a manually built ground truth. 
The results are evaluated according to different metrics.
We remark the role of the \Fmeasure, the harmonic average of precision and recall; 
and the Cohen's $\kappa$, which indicates the rate of agreement between the two classes.
We consider both of them as being high, in particular when compared to other segmentation tasks.

\begin{table*}[!ht]
    \centering
    \caption{Performance metrics using the \GradientBoosting and the \watershed algorithms.}
    \label{tab:results_rates}
    \begin{tabular}{@{}lcccccc@{}}
        \toprule
        \multicolumn{1}{l}{Dataset} & 
        \multirow{2}{0.12\textwidth}{\parbox{0.12\textwidth}{
            \centering \Sensitivity \\ {\scriptsize (average)}
        }} &  
        \multirow{2}{0.12\textwidth}{\parbox{0.12\textwidth}{
            \centering \Specificity \\ {\scriptsize (average)}
        }} &  
        \multirow{2}{0.12\textwidth}{\parbox{0.12\textwidth}{
            \centering \Precision \\ {\scriptsize (average)}
        }} &  
        \multirow{2}{0.12\textwidth}{\parbox{0.12\textwidth}{
            \centering \Accuracy \\ {\scriptsize (average)}
        }} &  
        \multirow{2}{0.12\textwidth}{\parbox{0.12\textwidth}{
            \centering \Fmeasure \\ {\scriptsize (average)}
        }} &  
        \multirow{2}{0.12\textwidth}{\parbox{0.12\textwidth}{
            \centering \CohensKappa \\ {\scriptsize (average)}
        }} 
        \\ \\ \midrule
        Train set      & 0.955 &  0.998 & 0.957 &     0.996 &        0.955 &      0.953 \\
        Test  set      & 0.925 &  0.997 & 0.936 &     0.993 &        0.925 &      0.922 \\ \bottomrule
    \end{tabular}
\end{table*}
\section{Conclusions}
\label{sec:Conclusion}

This paper introduces a nail segmentation for the big toe, based on a series of image processing operators.
Each of them is based on the information provided by the previous one.
%, as shown in Fig. \ref{fig:FlowDiagram}.
Far from amplifying the errors, subsequent steps are designed to correct possible deviations, refining the result provided by the previous step.

In the following, we discuss how different techniques exploit the different aspects of our task -- segmenting nails from the rest of the skin.
We end the paper with some remarks of the limitations and contributions that this work represents.

\subsection{Qualitative analysis}

In this section we discuss the rationale of how different steps take advantage of the problem characteristics.
% Conversely, we gather here a series of insights regarding what characteristics may be leveraged to segment nails from skin regions.

Nail and skin can not be segmented using only color or local information.
The pixel-wise colors of human nails are indistinguishable from those belonging to toes (see Sect.~\ref{sec:ToenailDetection}).
This is specially important when dealing with different skin shades or illumination conditions.
We can not, however, disregard photometric information (see Fig.~\ref{fig:bars_importances}).

Although nails present a diversity of shapes, they tend to fit well in a geometric circle.
However, such circle does not necessarily delimit its actual edges.
% Also, the toe end also fits in a circle, which is stronger in terms of its roundness and sharpness of contours (we emphasize that, in our case, the toe was located over a black background).
%To correctly detect the nail circle, we removed the contours outside the tip of the toe region so that the contours of the nail would have the highest magnitude.
The \Houghtransform was successfully estimate the nail circle after removing the between toe and background.
% These circles represented a very important piece of information for subsequent analysis (see Sect. \ref{sec:Results:Analysis}).

Along the same line, the boundary between nail and toe is really sharp but hard to discriminate from spurious contours.
% Thus, successfully leveraging this boundary is a hard task when processing images automatically.
Using the \watershed algorithm, we leverage the fact that nails are very well defined by their contours, even better than by their colorimetry, size or shape.
% In particular, choosing the \watershed markers appropriately is very important.
To use this boundary, we must know in advance a spatial location inside the nail and another one outside it.

\subsection{Discussion}

The algorithm introduced is designed to robustly segment the toenail from images taken from different cameras, angles and  lighting conditions. It is being currently used in the clinical practice.
In particular, it helps to measure the nail surface, easing the temporal analysis of patients suffering from a pathology that affects their toenail.

% Limitations
Two limitations must be mentioned.
First, we only considered skin shades from Caucasian patients due to the locations where the project was being held.
Second, a large contribution of the error is located in the lateral area of nails.
These darker regions tend to be wrongly estimated as skin.
However, this tends to have small effects on measurements of the nail area. 

% Good results
Despite such error-prone regions, we consider that our algorithm reliably segments nails from human toes, successfully addressing the task for which it was designed.
We consider so based on the qualitative examination of several samples, its robustness across the whole set of images, and on the quantitative metrics obtained.

% Scientific experimentation
A good experimental framework also substantiates our previous claim.
Performance measures have been obtained by considering disjoint training and test sets---although the test set was used to select the best classifier, which could have leaked some bits of information of the otherwise unseen test set.
Also, the black background, much easier to segment, has not been taken into account to compute the metrics. As the data belongs to a  pharmaceutical company  the database used for this study must remain private.

%XAI
Here, we want to remind that for trusting the behavior of intelligent systems, especially in medicine, they must be able to explain their decisions and actions to human users through techniques that produce more explainable models whilst maintaining high performance levels~\cite{adadi2018peeking}. For this reason, we analysed the importance of each feature used to distinguish nail and toe regions over the trees that conform the gradient boosting ensemble. Feature relevance techniques seem to be one of the most used scheme as a post-hoc explainability technique in the field of tree ensembles~\cite{arrieta2020explainable}. Then, we can conclude that the proposed method can be described as an explainable intelligent system.

%  Controlled template
Although the data acquisition was somewhat controlled with the use of a template, the photographs were captured with off-the-shelf smartphones, with uneven illumination and during clinical practice. 
% Hand nails -> Future work
The aforementioned robustness of the algorithm is a prerequisite to consider broadening its scope.
In this direction, its use could also be proposed (i) in less controlled environments and (ii) to segment hand nails, since is a very similar task to toe nail segmentation.

\section*{Declarations}

\subsection*{Funding}
We acknowledge the Ministerio de Economía, Industria y Competitividad (MINECO), the Agencia Estatal de Investigación (AEI), and the European Regional Development Funds (ERDF) for its support for the projects \mbox{TIN 2016-75404-P} (AEI/FEDER, UE), \mbox{TIN2016-81143-R} (AEI/FEDER, UE), and EXPLAINING (PID2019-104829RA-I00 / AEI / 10.13039/501100011033).

We also acknowledge the Govern de les Illes Balears for its support for the project PROCOE/2/2017. \mbox{P. Bibiloni} also benefited from the fellowship \mbox{FPI/1645/2014} under an operational program co-financed by the European Social Fund.

\subsection*{Conflicts of interest}
The  authors certify  that  they  have  NO  affiliations  with  or  involvement  in  any organization or entity with any financial interest (such as honoraria; educational grants; participation in speakers’ bureaus;  membership, employment, consultancies, stock ownership, or other equity interest; and expert testimony or patent-licensing arrangements), or non financial interest (such as personal or professional relationships, affiliations, knowledge or beliefs) in
the subject matter or materials discussed in this manuscript.

\subsection*{Availability of data and material}
The images used in this article belong  to the company that conducted the clinical study.
\subsection*{Code availability}

The source code used in this article belongs to the company that conducted the clinical study.

\subsection*{Acknowledgments}
We would also thank Syntax for Science, a company founded in Spain specialized in clinical research which provides biometrics-related services. APSL has been the technological provider of this project.

\bibliographystyle{apa}
\bibliography{main/references.bib}

\vspace{0.25cm}
\begin{minipage}[t][5cm][t]{0,9\textwidth}
\begin{wrapfigure}{i}{25mm} 
\includegraphics[width=0.2\textwidth,keepaspectratio]{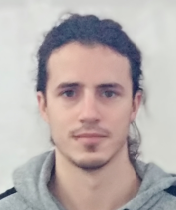}
\end{wrapfigure} 
\textbf{Bernat Galmés Rubert} is a PhD student at the Balearic Islands University (UIB). He obtained his Master’s degree in computer science in 2018 at the Open University of Catalonia.
\end{minipage}

\begin{minipage}[t][5cm][t]{0,9\textwidth}
\begin{wrapfigure}{l}{25mm} 
\includegraphics[width=0.2\textwidth,keepaspectratio]{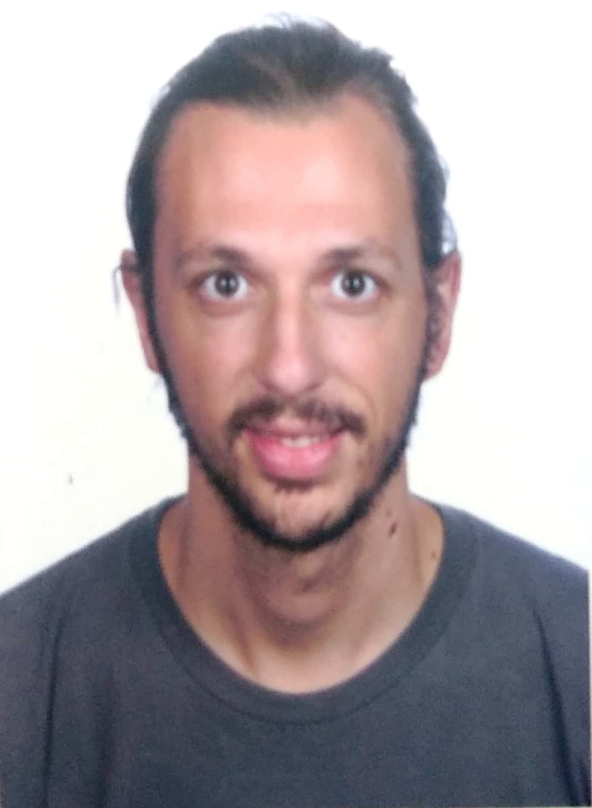}
\end{wrapfigure} 
\textbf{Gabriel Moyà Alcover} obtained the Ph.D. in Computer Science in 2016. His main research interests are mainly about computer vision.  He is researcher at the Computer Graphics and Vision and Artificial Intelligence Group.
\end{minipage}

\begin{minipage}[t][5cm][t]{0,9\textwidth}
\begin{wrapfigure}{l}{25mm} 
\includegraphics[width=0.2\textwidth,keepaspectratio]{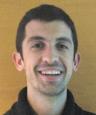}
\end{wrapfigure} 
\textbf{Pedro Bibiloni} is a lecturer at the University of the Balearic Islands. He received a Ph.D. at the University of the Balearic Islands (2018). His research interests are medical image processing and pattern recognition.
\end{minipage}

\begin{minipage}[t][5cm][t]{0,9\textwidth}
\begin{wrapfigure}{l}{25mm} 
\includegraphics[width=0.2\textwidth,keepaspectratio]{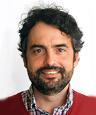}
\end{wrapfigure} 
\textbf{Javier Varona Gómez} PhD in Computer Science. He has participated in over 10 research funded projects (in 5 of them as principal investigator). Among his research interests are computer vision and artificial intelligence. 

\end{minipage}

\begin{minipage}[t][5cm][t]{0,9\textwidth}
\begin{wrapfigure}{l}{25mm} 
\includegraphics[width=0.2\textwidth,keepaspectratio]{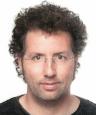}
\end{wrapfigure} 
\textbf{Antoni Jaume-i-Cap\'o}: PhD in Computer Science. Among his research lines are computer vision, medical imaging processing. He is researcher at the Computer Graphics and Vision and Artificial Intelligence Group.
\end{minipage}

\end{document}